\documentclass{article}

\usepackage{arxiv}

\usepackage[utf8]{inputenc} 
\usepackage[T1]{fontenc}    
\usepackage{hyperref}       
\usepackage{url}            
\usepackage{booktabs}       
\usepackage{amsfonts}       
\usepackage{nicefrac}       
\usepackage{microtype}      
\usepackage{lipsum}		
\usepackage{graphicx}
\usepackage{subcaption}
\usepackage{mathrsfs}
\usepackage{amsmath}

\title{Predicting heave and surge motions of a semi-submersible with  neural networks}

\author{ {Xiaoxian Guo} {, Xiantao Zhang, } {Xinliang Tian, } {Xin Li, } {Wenyue Lu}\\
		State Key Laboratory of Ocean Engineering\\
		Shanghai Jiao Tong University\\
		Shanghai 200240, China \\
	}

\begin{document}
\maketitle

\begin{abstract}
	Real-time motion prediction of a vessel or a floating platform can help to improve the performance of motion compensation systems. It can also provide useful early-warning information for offshore operations that are critical with regard to motion. In this study, a long short-term memory (LSTM) -based machine learning model was developed to predict heave and surge motions of a semi-submersible. The training and test data came from a model test carried out in the deep-water ocean basin, at Shanghai Jiao Tong University, China. The motion and measured waves were fed into LSTM cells and then went through serval fully connected (FC) layers to obtain the prediction. With the help of measured waves, the prediction extended 46.5 s into future with an average accuracy close to 90\%. Using a noise-extended dataset, the trained model effectively worked with a noise level up to 0.8. As a further step, the model could predict motions only based on the motion itself. Based on sensitive studies on the architectures of the model, guidelines for the construction of the machine learning model are proposed. The proposed LSTM model shows a strong ability to predict vessel wave-excited motions.
\end{abstract}

\keywords{Semi-submersible \and Motion prediction \and Wave-excited motion \and Neural network \and LSTM}

\section{Introduction}
Offshore human activities  such as oil and gas exploration, deep sea expedition,  and military missions rapidly advanced in the past decades. In deep waters (water depth greater than 500 m), most of the offshore platforms float, which means that they can move in six degrees of freedom (DOFs) excited by environmental forces, such as waves, winds, or ocean currents. The 6-DOF motions of the platform constitute a remarkable challenge to carry out works offshore, thereby limiting the ability of many operations that are critical with regard to motion. Much of the equipment requires to work with a motion compensation system in a harsh sea state, e.g., a dynamic gangway to transfer crews from ship to ship. Real-time (also called online) motion prediction of a vessel or a floating platform usually refers to forecasting vessel motions in the following tens of seconds (one or two wave cycles) in real time. It can help improve the performance of a motion compensation system. It can also provide useful early-warning information. An aircraft landing on a carrier or a missile launched from a warship also need real-time motion prediction for better performance. 

The 6-DOF motions of an offshore platform are caused by environmental excitations and are restricted by mooring systems to keep the position . Such motions can be divided into two groups. First, for heave, roll, and pitch motions, water provides restoring force; the motions in these directions only have wave frequency components. Second, for surge, sway, and yaw motions, the mooring lines provide restoring force to keep the position. The natural frequencies are much lower than the wave excitation frequency in these directions. Therefore, the surge motion of a platform usually consists of the wave frequency and low-frequency (slow drift) parts~\cite{Faltinsen1993}. 

The dynamic responses of offshore platforms were thoroughly studied. Although environmental excitations in real seas are random, we believe that the process is stationary and ergodic~\cite{Ochi1990}. In traditional calculations, the motion responses are usually described by response spectra or response amplitude operators (RAOs) in the frequency domain, and statistical values such as mean, standard deviation, and significant values in the time domain. Such analysis is performed in frequency or time domain based on time series from 3-hour simulations or experiments~\cite{Zhao2014,Qi2019,Dong2019}. Thus, the exact platform motion is not described in one or two wave cycles; statistical values and spectra are used as key features to describe the motion in a 3-hour period. These results contribute to the safety design of offshore platforms.

Triantafyllou et al.~\cite{Triantafyllou1981,Triantafyllou1983} predicted 6-DOF ship motions with an upper bound of approximately 5 s using Kalman filters. The Kalman filtering technique can efficiently work with noisy measurements but requires the state-space model of the vessel based on full knowledge of hydrodynamics. Given that the added mass, damping, and wave excitation forces are frequency-dependent, the estimation of the peak frequency of the wave spectrum is crucial, and therefore limits its application in real seas. Recently, some systems for real-time prediction of vessel motions were developed~\cite{Naaijen2009, Alford2015}. They were based on nonlinear wave and vessel seakeeping theories. All the above models need full knowledge of hydrodynamics. Yumori~\cite{Yumori1981} demonstrated an auto-regressive moving average  model to predict the ship behavior approximately 2 to 4 s ahead. Direct time-series analysis does not require prior knowledge about the ship responses. Broome et al.~\cite{Broome1990} compared different mathematical auto-regressive (AR) models for online motion prediction. It was found that AR models are valid for predicting most of the roll time series but cannot predict sudden bursts. 

Machine learning (ML) offers a powerful information framework to extract features directly from labeled data without deep knowledge of the underlying physics. The unknown parameters in the model are learned by minimizing the loss between prediction and ground truth through a back propagation algorithm. Sclavounos and Ma~\cite{Sclavounos2018} used support vector machines  to perform a short-term wave elevation forecast that could predict 5 s into future with good accuracy. Neural networks (NNs) are the most popular methods among ML algorithms. Khan et al.~\cite{Khan2005} achieved roll motion prediction extending up to 7 s using a 3-layer FC NN model. Li et al.~\cite{Li2019} used NNs with 2 hidden layers to predict a wave excitation force extending 2.5 s ahead for the controller of a wave energy converter. Deep neural networks (DNNs), which are NNs with many hidden layers, could be optimal nonlinear approximators for a wide range of functions~\cite{Bolcskei2019}. It seems that the prediction performance can be easily improved by using deeper and wider NNs. The only aspect to consider is the trade-off between the model accuracy
and the computation cost. However, it is a fact that a larger parameter space needs a very large amount of data to train the model. DNNs also suffer from over-fitting, i.e., the learned model is only valid for some specific scenarios. Another problem is that the input and output sizes of a FC NN are fixed. The feeding data length and predicting time steps cannot be changed.

To address sequential data, recurrent neural networks (RNNs) were designed. These networks consider the inherent order of the data and are not subjected  to the input data length. RNNs are enormously successful in natural language processing. At present, so-called long short-term memory (LSTM) \cite{Hochreiter1997} algorithms are the most popular type of RNN. They can help to overcome diminishing or exploding gradients during the training process. LSTMs can predict time series accurately for many different tasks, such as traffic and user mobility~\cite{Hua2019}, stock prices~\cite{Nelson2017}, and weather conditions~\cite{Karevan2020}. LSTM models also showed their feasibility for ship motion prediction~\cite{Hua2019,Duan2019}. Ferrandis et al.~\cite{Ferrandis2019} compared a simple RNN, a gated recurrent unit, and LSTM algorithms for ship motion prediction. They concluded that LSTM models achieve the best performance.

The objective of this study was to carry out a multi-step prediction on heave and surge motions of a semi-submersible up to 50 s ahead in real time using an LSTM model. A model test was performed in a deep-water ocean basin to obtain training and test data. It should be pointed out that most of the considered offshore platforms were tested in the wave basin during the design stage. Thus, if the training process provided successful predictions based on the existing model test data, the trained model could be used on real offshore platforms without large-scale field measurements. This paper is organized as follows. In the next section, we describe the basic idea underlying the proposed ML algorithm and the generation of the labeled data from model test results. In Section 3, three examples that were carried out as a benchmark are described. We first predicted heave and surge motions with the help of wave elevation measurements. Thereafter, the model was trained to address noisy inputs. Finally, the predictions were only based on the motion itself. A sensitive study on the hyperparameters, model structure, and feeding data length helps to understand the physics underlying the ML model and gives general guidelines for real-time motion prediction of floating offshore platforms.

\section{Methodology}
\subsection{Overview of the procedure} \label{sec:method}
The basic idea of a ML algorithm can be described as follows. For a pair of input tensor $\mathbf{x} \in \mathbb{R}^{r \times n}$ and output tensor $\mathbf{y} \in \mathbb{R}^m$, we have to find the parameters $\mathbf{p}$ (weights and bias) in a learning machine $\mathscr{F}$ that performs nonlinear mapping expressed as follows:
\begin{equation}
	\mathscr{F}(\mathbf{x}; \mathbf{p})\approx \mathbf{y};
\end{equation}
in this study, the input tensor consists of sequential input features such as wave elevation and motions, and the output tensor provides the predicted motions.

The loss function is defined as mean square error (MSE):
\begin{equation}
	\mathscr{L} = \sum_{i=1}^{N} \frac{\big|y_i - \mathscr{F}(\mathbf{x}; \mathbf{p})_i\big|^2}{N}.
\end{equation}

\begin{figure}[ht]
	\centering
	\includegraphics[width=0.9\linewidth]{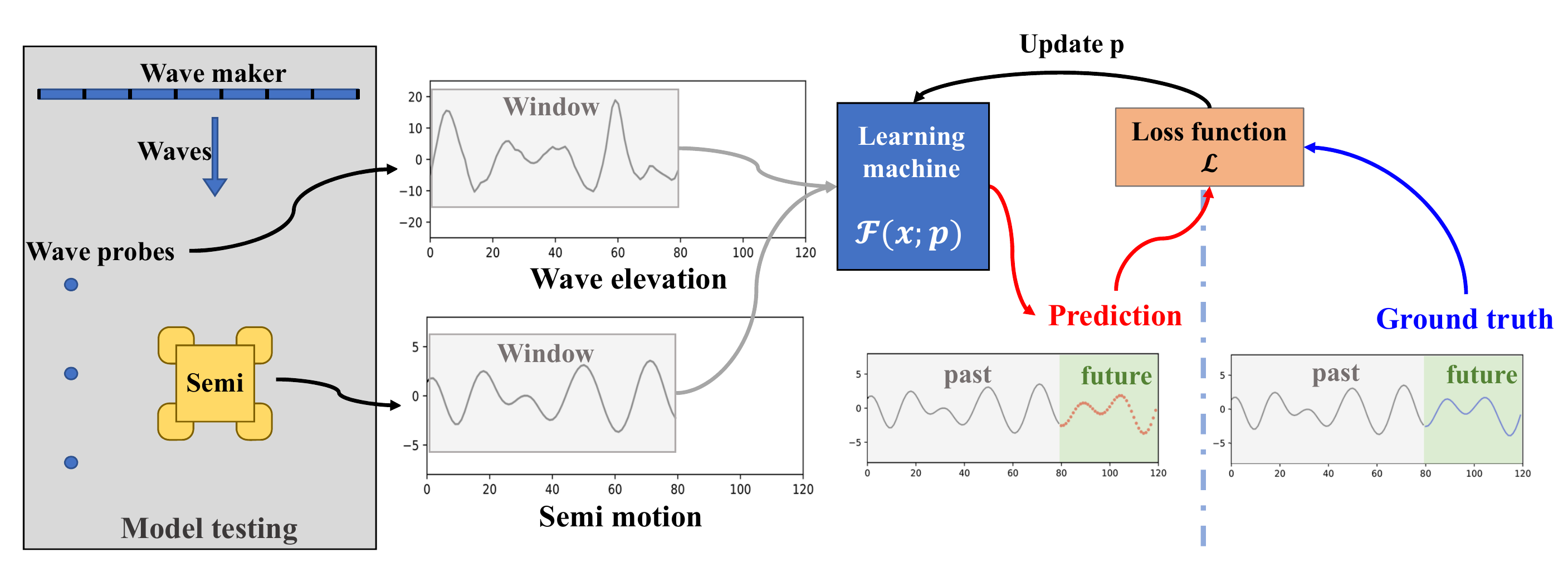}
	\caption{An overview of the training process of the learning machine for real-time motion prediction.}
	\label{fig:overall}
\end{figure}

The overall procedure of the real-time motion prediction learned by the machine is presented in Fig.~\ref{fig:overall}. All the training and test data came from a model test in a wave basin. The measured time series of wave elevation and the model motion were fed into the learning machine to generate the prediction. The obtained prediction was compared with the ground truth through the loss function. By minimizing this loss, the parameters in the learning machine were updated. These steps were repeated until sufficient prediction accuracy was obtained. Once the parameters (weights and bias) were learned, the ML model $\mathscr{F}$ was ready to use.

\subsection{LSTM neural network}
\begin{figure}
	\centering
	\begin{subfigure}{\linewidth}
         \centering
         \includegraphics[height=4.5cm]{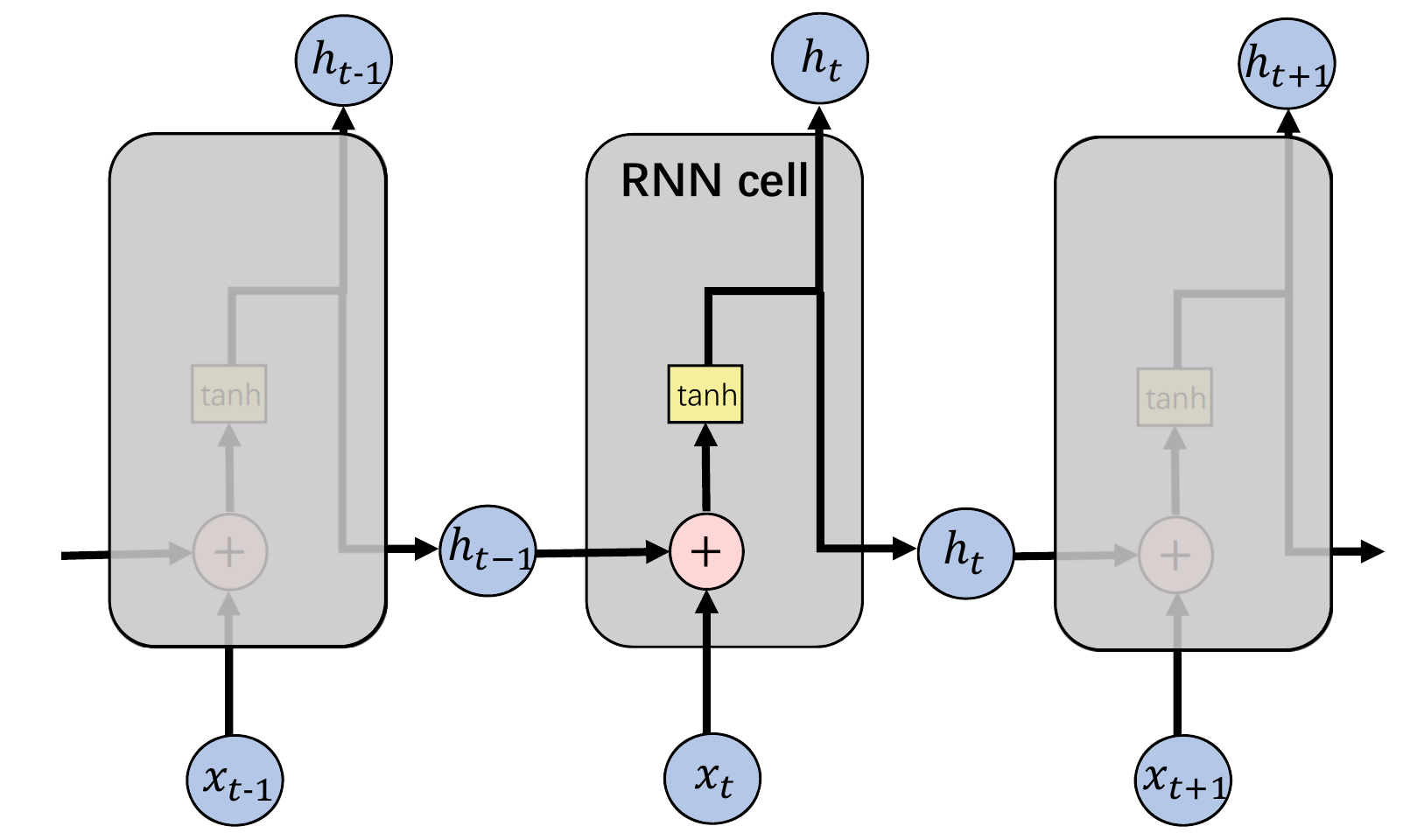}
         \caption{rnn cell}
         \label{fig:cells:rnn}
	\end{subfigure}
	\hfill
	\begin{subfigure}{\linewidth}
		\centering
		\includegraphics[height=4.5cm]{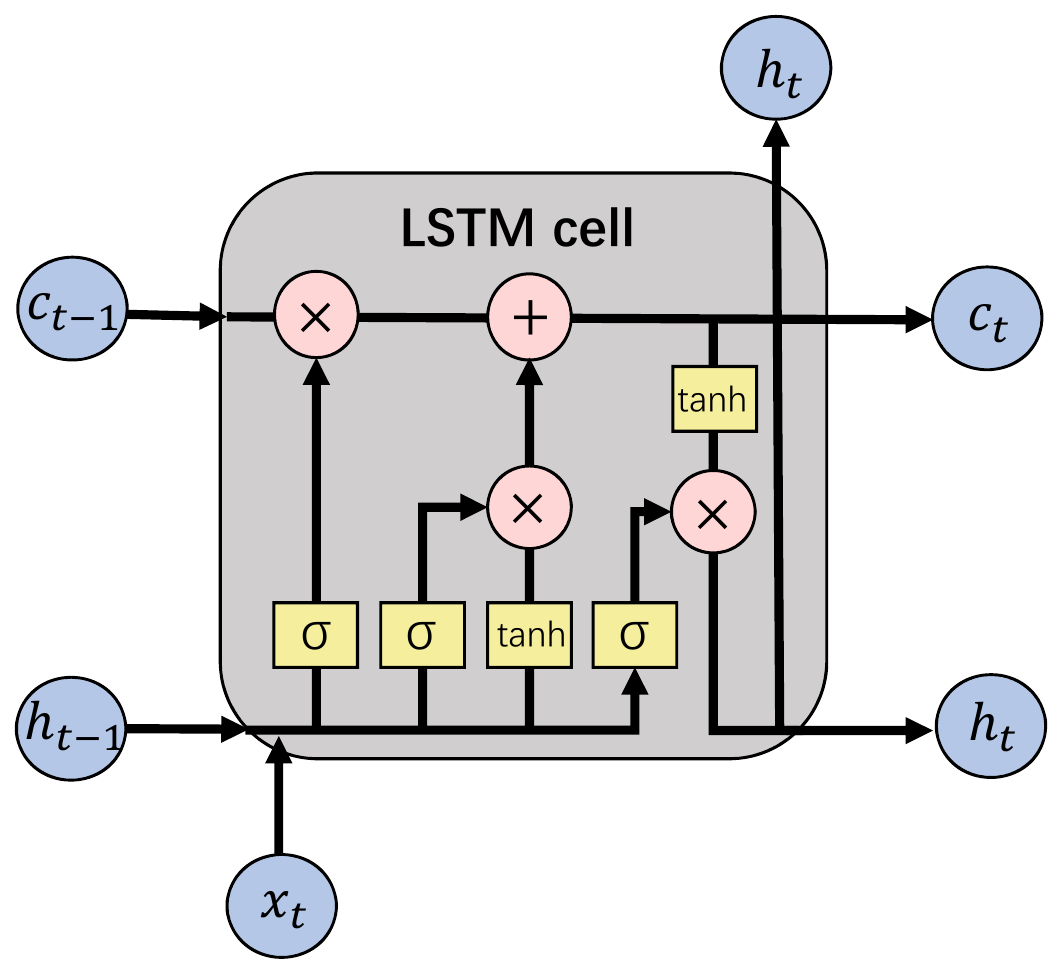}
		\caption{LSTM cell}
		\label{fig:cells:lstm}
    \end{subfigure}
	\caption{Basic architectures of (a) an RNN cell and (b) an LSTM cell.}
	\label{fig:cells}
\end{figure}

An RNN is a generalization of feed-forward neural network with an internal memory. As shown in Fig.~\ref{fig:cells} (a), the basic idea is that the parameters in each cell are the same, and the output at the previous time step feeds the current cell as memory. The parameter size is independent of the length of the sequential input data, and with the memory flowing through the cells, the RNN network provides a better performance for sequential input data. However, with increasing length of the input sequence, the memory goes through the activation function too many times, which may lead to gradient vanishing problems. The basic RNN provides a good basis to address sequential data, but its training for real tasks is very difficult. An LSTM model is a modified version of RNN. As shown in Fig.~\ref{fig:cells} (b), the memory can easily spread through the cells by adding $c_{t}$, which acts as a shortcut between the cells. LSTMs are well-suited to classify, process, and predict time series given time lags of unknown duration. The formulation of an LSTM cell is as follows~\cite{Hochreiter1997}:

\begin{align}
&i_{t}=\sigma\left(W_{i i} x_{t}+b_{i i}+W_{h i} h_{t-1}+b_{h i}\right) &\text{(input gate)}\\
&f_{t}=\sigma\left(W_{i f} x_{t}+b_{i f}+W_{h f} h_{t-1}+b_{h f}\right) &\text{(forget gate)}\\
&g_{t}=\tanh \left(W_{i g} x_{t}+b_{i g}+W_{h g} h_{t-1}+b_{h g}\right) &\text{(cell gate)}\\
&o_{t}=\sigma\left(W_{i o} x_{t}+b_{i o}+W_{h o} h_{t-1}+b_{h o}\right) &\text{(output gate)}\\
&c_{t}=f_{t} \odot c_{t-1}+i_{t} \odot g_{t} &\text{(cell state)}\\
&h_{t}=o_{t} \odot \tanh \left(c_{t}\right) &\text{(hidden state)}
\end{align}
where $W$ and $b$ are the weights and bias, respectively, $\sigma$ is the sigmoid function, and $\odot$ represents element-wise multiplication.

\subsection{Model test}
\begin{figure}[ht]
	\centering
	\includegraphics[height=6.5cm]{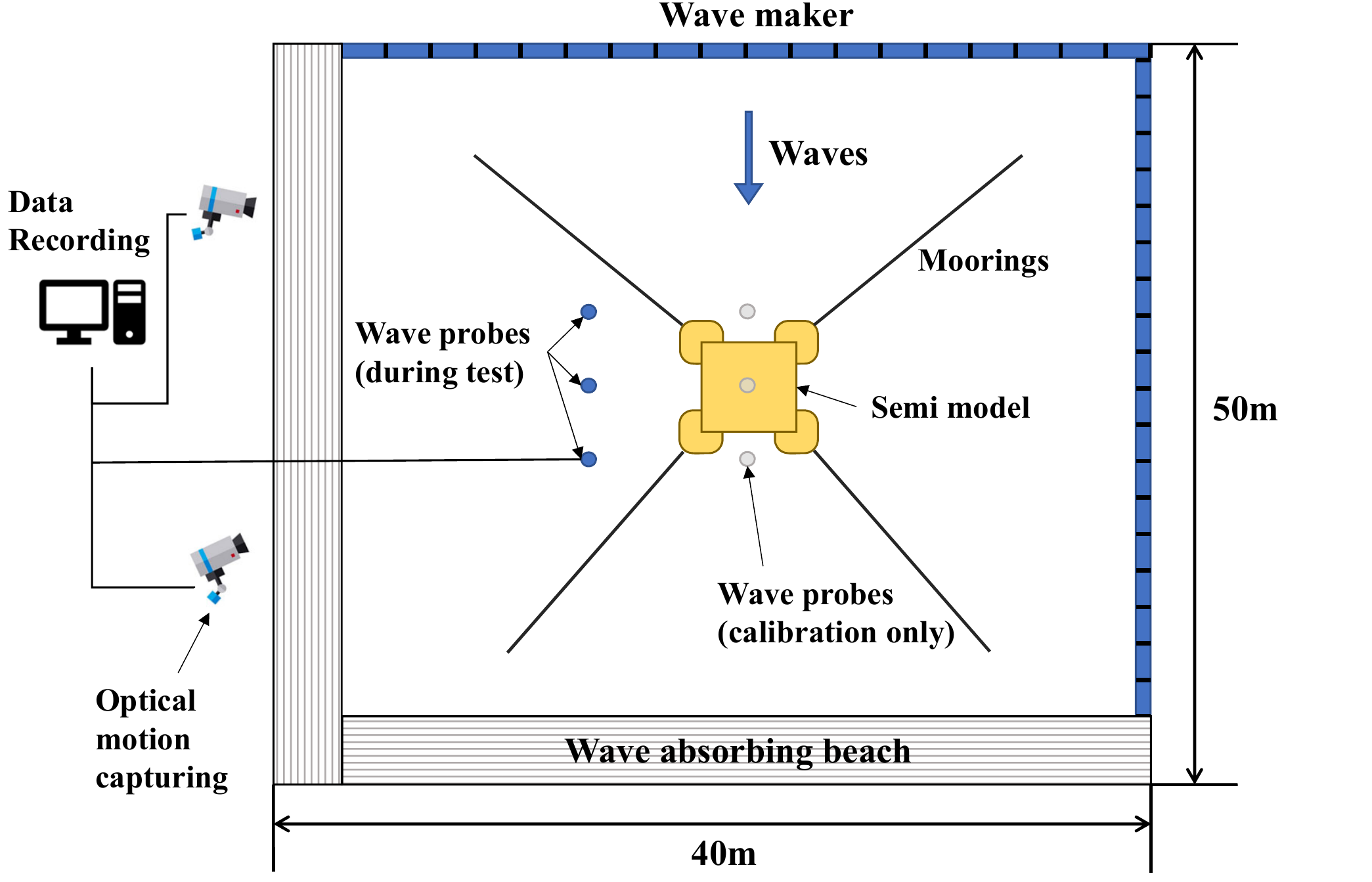}
	\caption{Top view of the semi-model arrangement in the deep-water wave basin.}
	\label{fig:basinview}
\end{figure}

A scaled model test was performed in this study to obtain the training and test data in the deep-water wave basin at Shanghai Jiao Tong University (SJTU), China. This wave basin (see Fig.~\ref{fig:basinview}) is 50.0 m in length, 40.0 m in width and up to 10.0 m in depth with a large-area movable bottom. The L-shaped wave maker, which spans two adjacent sides, was equipped to model the ocean waves in the basin. The passive wave-absorbing beach was also equipped with an optimized parabola profile and damping grids for wave dissipation. 

\begin{figure}[ht]
	\centering
	\begin{subfigure}{0.4\linewidth}
         \centering
         \includegraphics[width=\linewidth]{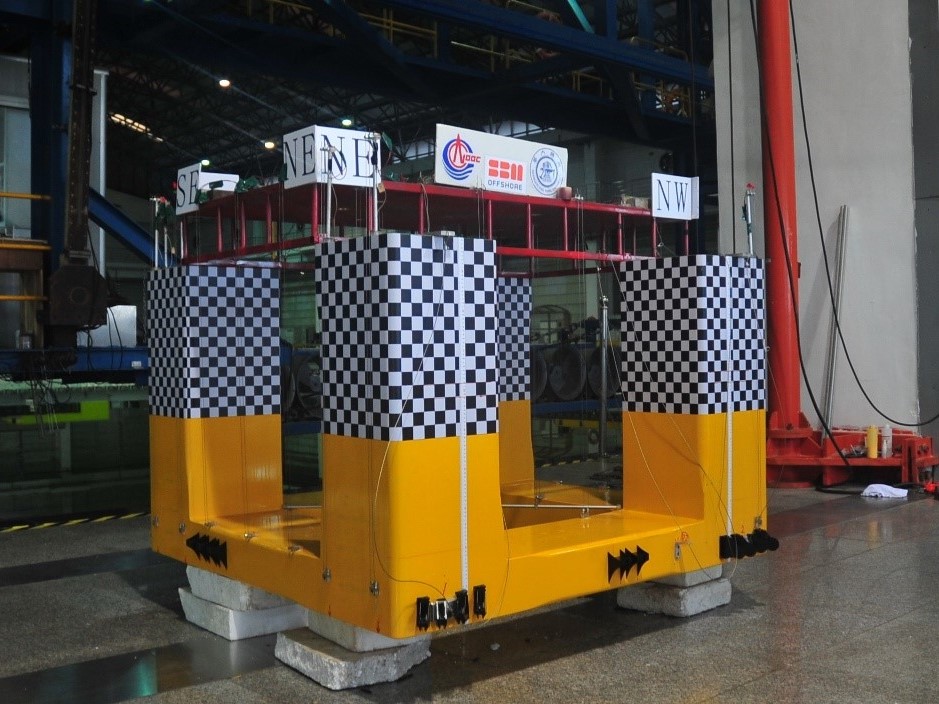}
         \caption{Semi-submersible model (1:60).}
         \label{fig:model:pic}
	\end{subfigure}
	\hfill
	\begin{subfigure}{0.4\linewidth}
		\centering
		\includegraphics[width=\linewidth]{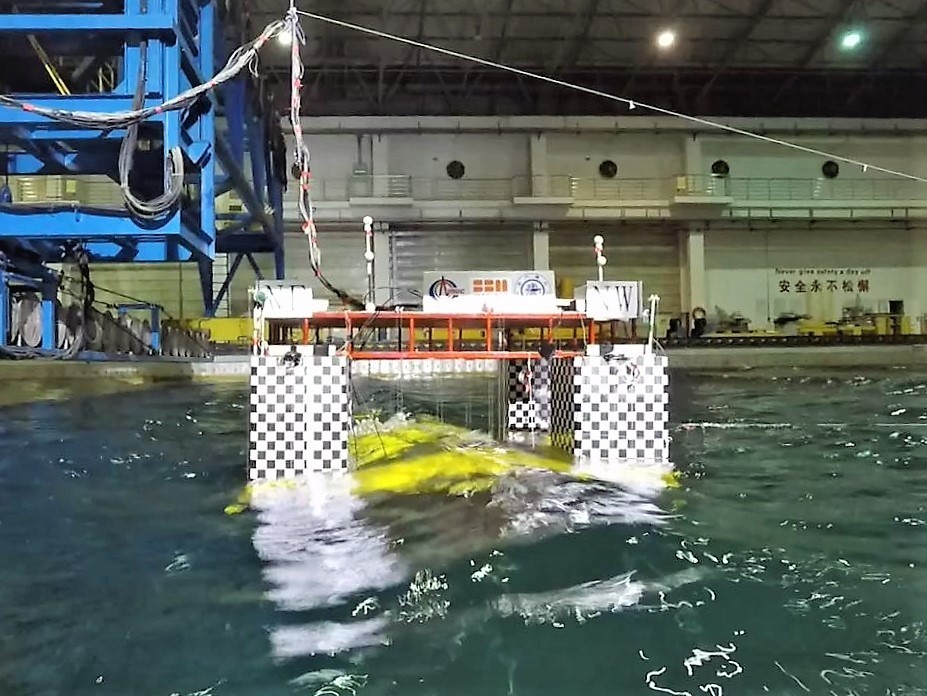}
		\caption{The semi-submersible model was tested in water.}
		\label{fig:model:inwater}
    \end{subfigure}
	\caption{Pictures of (a) semi-submersible model, and (b) in-water model employed during the experiments.}
	\label{fig:model}
\end{figure}

The scaled model, shown in Fig.~\ref{fig:model} (a), is a typical semi-submersible platform employed for offshore oil and gas industry. The scale ratio was set to 1:60. The main dimensions and properties of this semi-submersible platform are listed in Tab.~\ref{tab:miandim}. The model was made by wood and adjusted on the trimming table to ensure its correct properties, including mass, center of gravity, and radius of gyration. Fig.~\ref{fig:basinview} shows that the model was moored in the wave basin by 4 catenary mooring lines attached to the model at corners. The catenary mooring system provided restoring force. Therefore, it helped the model to keep its position in harsh sea state. An optical motion capture system was used for capturing the 6-DoF motions at the center of the waterline area of the model. Three wave probes were installed beside the model to record the wave elevations. A detailed description of the model test can also be found in \cite{Li2020}.

\begin{table}[ht]
	\caption{Main dimensions and properties of the semi-submersible platform.}\label{tab:miandim}
	\centering
	\begin{tabular}{lccc}
		\toprule
		Parameter & Unit  & prototype & model \\
		\midrule
		Normal Draft & m     & 37.0  & 0.617 \\
		Hull Width & m     & 91.5  & 1.525 \\
		Column spacing C/C & m     & 70.5  & 1.175 \\
		Pontoon width & m     & 21.0  & 0.35 \\
		Pontoon height & m     & 9.0   & 0.15 \\
		Nominal Displacement & MT    & 105000  & 474.2kg \\
		KG    & m     & 32.0  & 0.533 \\
		Roll gyradius & m     & 39.0  & 0.65 \\
		Pitch gyradius  & m     & 41.0  & 0.683 \\
		Yaw gyradius & m     & 42.0  & 0.7 \\
		\bottomrule
	\end{tabular}
\end{table}

The generated irregular waves followed the JONSWAP  spectrum~\cite{DNVRP2010} (see Equ.~(\ref{equ:jonswap})). All the considered wave conditions, with varying significant wave height and wave period, are listed in Tab.~\ref{tab:wavecond}. In total, 8 different wave conditions were considered, corresponding to 100-year and 1000-year cyclones.

\begin{equation}\label{equ:jonswap}
  S(\omega)= \alpha H_s^2 \frac{\omega^{-5}}{\omega_p^{-4}} \exp\big[-1.25(\omega/\omega_p)^{-4}\big]
  \gamma^{\exp\big[-\frac{(\omega-\omega_p)^2}{2\tau^2\omega_p^2}\big]},
\end{equation}
where $\gamma$ is the so-called peakedness parameter ($\gamma=2.4$ throughout this study), $\tau$ is the shape parameter ($\tau=0.07$ for $\omega>\omega_p$ and $\tau=0.09$ for $\omega<\omega_p$, as suggested in~\cite{DNVRP2010}), $H_s$ is the significant wave height, and $\omega_p=2\pi/T_p$ is the angular spectral peak frequency. 

\begin{table}[ht]
	\caption{Wave parameters used in the test prototype.}\label{tab:wavecond}
	\centering
	\begin{tabular}{llccll}
		\toprule
		No.   & Description & $H_s$ (m) & $T_p$ (s) & Note & Dataset\\
		\midrule
		WC1   & 100 yr cyclone & 13.4  & 14.2  & Short Tp & Training\\
		WC2   & 100 yr cyclone & 13.4  & 14.7  & Seed 1 & Test \\
		WC3   & 100 yr cyclone & 13.4  & 14.7  & Seed 2 & Training\\
		WC4   & 100 yr cyclone & 13.4  & 15.7  &  Long Tp& Training\\
		WC5   & 1000 yr cyclone & 16.9  & 14.4  & Short Tp & Training\\
		WC6   & 1000 yr cyclone & 16.9  & 15.9  & Seed 1& Training\\
		WC7   & 1000 yr cyclone & 16.9  & 15.9  & Seed 2 & Training\\
		WC8   & 1000 yr cyclone & 16.9  & 16.9  & Long Tp& Training \\
		\bottomrule
	\end{tabular}
\end{table}

\begin{figure}[ht]
	\centering
	\includegraphics[width=0.9\linewidth]{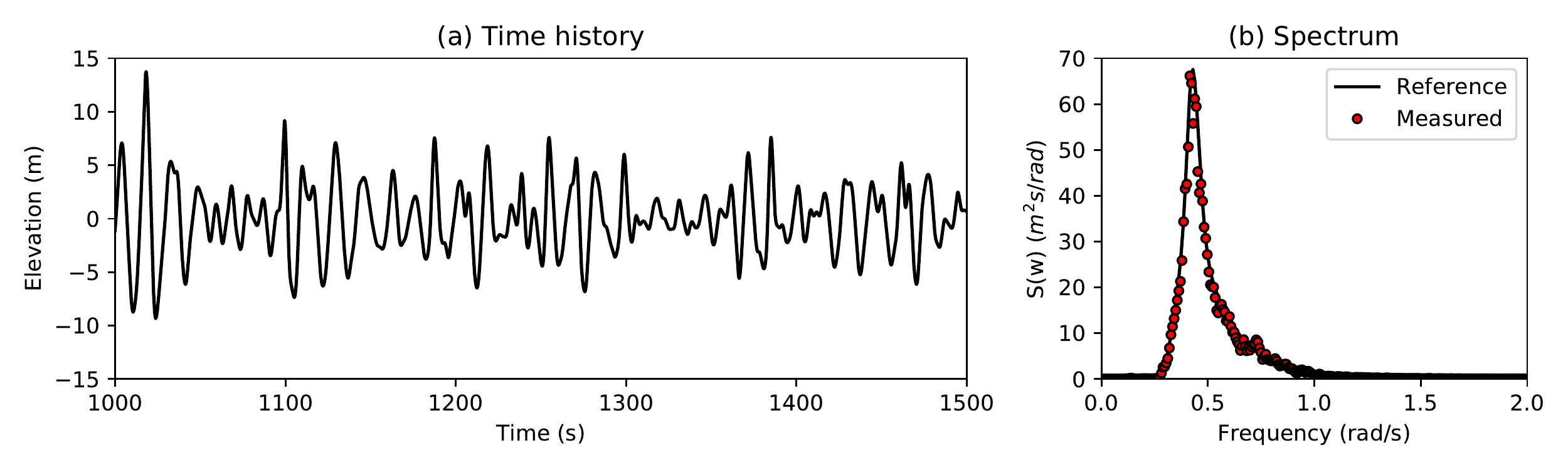}
	\caption{Wave calibration results of (a) time history and (b) spectrum for WC3 with $H_s=13.4$ m and $T_p=14.7$ s.}
	\label{fig:wavecalib}
\end{figure}

The specified wave conditions were calibrated before starting the model test. Instead of the model, three wave probes were placed at the center of the basin to record the wave elevation for wave calibration. Figure~\ref{fig:wavecalib} shows, as an example, the time series of the wave elevation at the center of the basin and corresponding spectrum for WC3 compared with theoretical results. 

After wave calibration, the model was put into the basin and moored by the proposed catenary mooring system. For each case, we generated waves and synchronously recorded the motions and wave elevation for 30 min (corresponding to 3 h in prototype). The sampling rate was set at 10 Hz in the model scale. Figure~\ref{fig:model} (b) shows the in-place model during testing.

\subsection{Training and test datasets}
\begin{figure}[ht]
	\centering
	\includegraphics[width=0.9\linewidth]{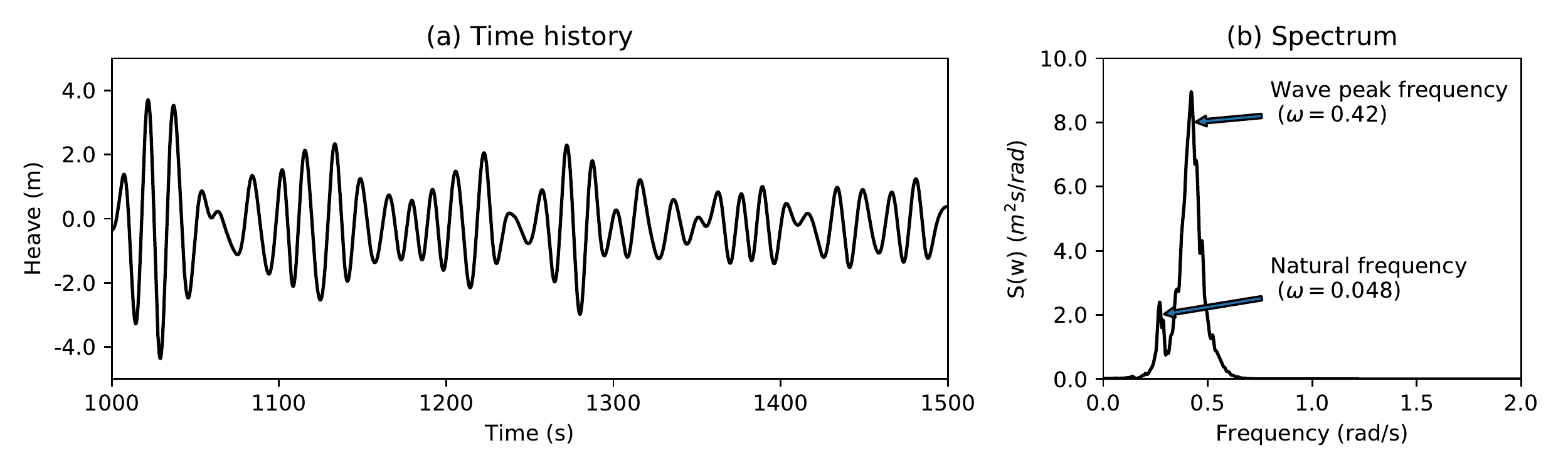}
	\caption{(a) Time history and (b) spectrum of platform heave motion under WC3 with $H_s=13.4$ m and $T_p=14.7$ s.}
	\label{fig:heave}
\end{figure}

\begin{figure}[ht]
	\centering
	\includegraphics[width=0.9\linewidth]{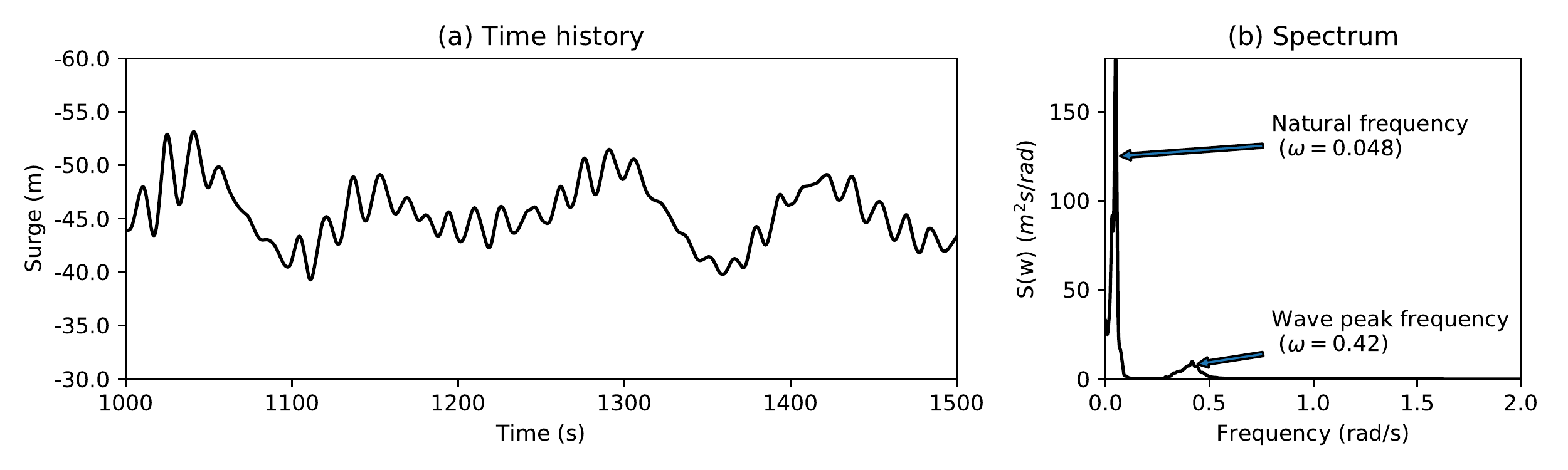}
	\caption{(a) Time history and (b) spectrum of platform surge motion WC3 with $H_s=13.4$m and $T_p=14.7$s.}
	\label{fig:surge}
\end{figure}

For each case, we synchronously recorded the wave elevation and 6-DOF motions of the model. Figures~\ref{fig:heave} and~\ref{fig:surge} show the time series and fast Fourier transform  spectra of heave and surge motions, respectively. Note that all the results were scaled up to a prototype with scale ratio of 60. For heave motion, the oscillation frequency equaled the incident wave peak frequency. Although the response peak at natural frequency can be identified in the response spectrum, most of the energy concentrates on the wave peak frequency motion. This is because the restoring force mainly came from buoyancy, and the vessel moved with the incident waves. For surge motion, the situation changes. Given that the mooring system provided the restoring force, the natural frequency was much lower than the wave frequency. However, at such low-frequency region, the damping was very small. Therefore, the amplitude of the low-frequency motion was much larger than that of the wave frequency motion. Clearly, the surge motion had two parts: wave frequency and low frequency. 

At this point, we had several 30-min time series of recorded motion $\{x(t_i), i\in N\}$ and wave $\{w(t_i), i\in N\}$. We prepared the data as input-output ($\mathbf{X}$-$\mathbf{Y}$) pairs as follows:
\begin{align}
	&\mathbf{X}^{n \times 2}_p = [x(t_{p-n}), x(t_{p-n}), \cdots, x(t_{p-1})]^T+[w(t_{p+w-n}), w(t_{p+w-n+1}), \cdots, w(t_{p+w+1})]^T \\
	&\mathbf{Y}^{m \times 1}_p = [x(t_{p}), x(t_{p+1}), \cdots, x(t_{p+m-1})]^T
\end{align}
where $t_p$ denotes the current time instant, $n$ is the time window expressed as the number of points of sequential wave and motion data used as inputs, $w$ is the wave lag expressed as the number of points of waves into future, and $m$ is the prediction length expressed as the length of the prediction into future. Here, the wave lag $w$ means that we know the number of points ahead. In practice, we usually measure the waves in front of the platform. The wave lag represents how far the wave probe was placed in front of the platform. 

\begin{figure}[ht]
	\centering
	\includegraphics[width=\linewidth]{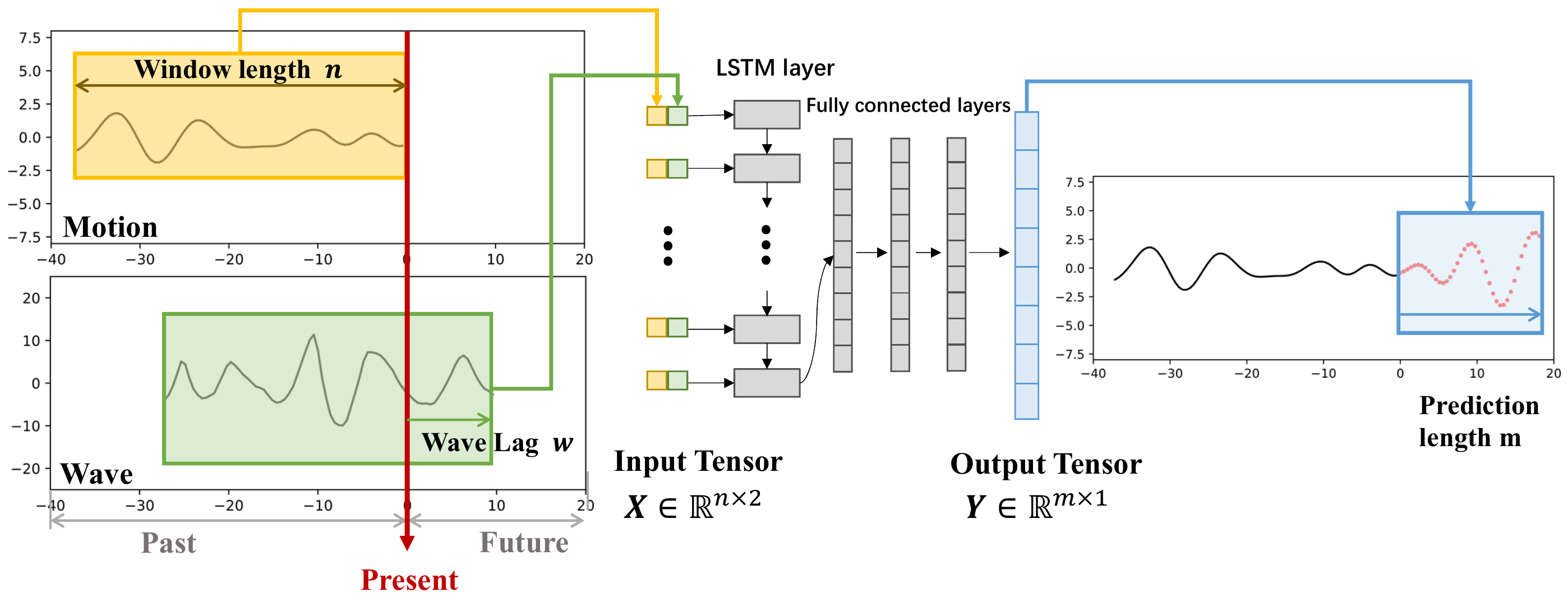}
	\caption{Illustration of the construction of the dataset.}
	\label{fig:dataset}
\end{figure}

To reduce the variance of the parameters in the NN, both the input waves and motions were regularized as follows:
\begin{equation}
	\widetilde{x}(t_i) = \frac{x(t_i)-\text{A}}{\text{B}},
\end{equation}
where A and B are the averaged mean and standard deviation of all the 30-min time series, respectively, which are constant in the present study. The actual values of A and B used are listed in Tab.~\ref{tab:ab}. One 30-min time history was first regularized, and then it was scaled up to full scale. Finally, it was divided into input-output tensor pairs as illustrated in Fig.~\ref{fig:dataset}.

\begin{table}[ht]
	\caption{Actual values of A and B in the present model. (Unit: cm)}\label{tab:ab}
	\centering
	\begin{tabular}{cccc}
		\toprule
		 & Heave & Surge & Wave\\
		\midrule
		A     & -0.86 & -100.341 & 0.422 \\
		B     & 2.264 & 7.876 & 6.766 \\
		\bottomrule
	\end{tabular}
\end{table}

\subsection{Training process}
The basic structure of the NN is shown in Fig.~\ref{fig:dataset}. The input data are first fed into LSTM cells. The output tensor of the last LSTM cell goes through three FC layers with hyperbolic tangent (tanh) as the activation function. Finally, the output tensor is obtained. The parameters of the basic configuration of the prediction model are listed in Tab.~\ref{tab:basicmodel}. The number of trainable parameters is 19470. 

\begin{table}[ht]
	\caption{Network structure of the motion prediction model.}\label{tab:basicmodel}
	\centering
	\begin{tabular}{ccc}
		\toprule
		No. & layer & Data size \\
		\midrule
		1 & Input layer & (256, n, 2) \\
		2 & LSTM layer & (256, n, 50, 1) \\
		3 & FC layer-1 & (256, 50, 1) \\
		4 & FC layer-2 & (256, 50, 1) \\
		5 & FC layer-3 & (256, 50, 1) \\
		6 & Output layer & (256, m, 1) \\
		\bottomrule
	\end{tabular}
\end{table}

\begin{figure}[ht]
	\centering
	\includegraphics[width=0.65\linewidth]{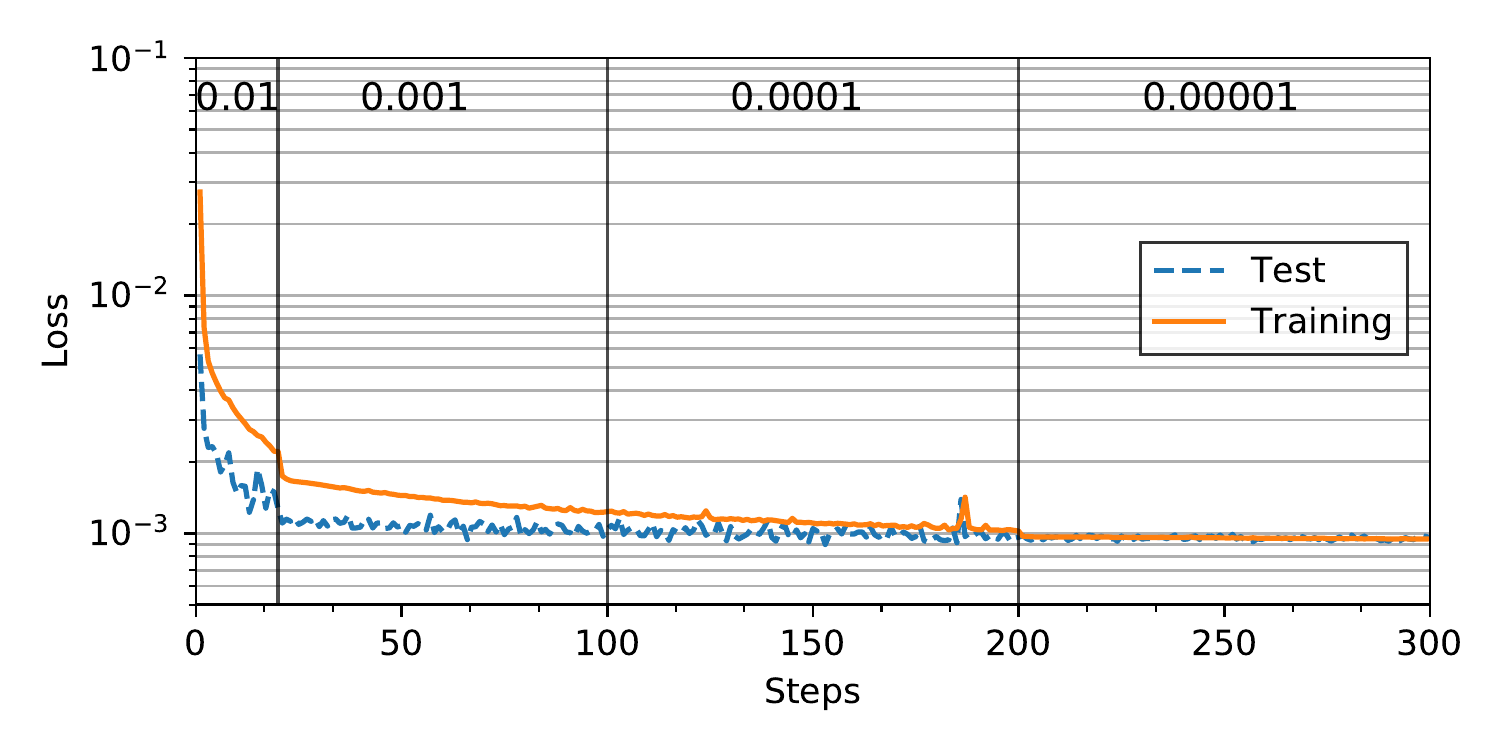}
	\caption{Training and test loss during the training process.}
	\label{fig:loss}
\end{figure}

In total, we had eight 30-min time series with different incident waves. For each series, we had approximately 15000 X-Y pairs, depending on the time window and prediction length. To ensure that the test data were completely unseen, we used data from WC1 and from WC3 to WC8 for training, leaving the WC2 case for test. 

Adam algorithm~\cite{Kingma2014} was used for minimizing the loss. The learning rate was set at 0.01 for the first 20 epochs, and then it was forced to decay at a rate of one tenth every 100 epochs. Mini-batch gradient descent was also applied with a batch size of 512. As an example, the loss of one training process is shown in Fig.~\ref{fig:loss}. 

\section{Results and discussion}
\subsection{Example 1: prediction with the help of wave measurements}
The motions of the semi-submersible platform were excited by incident waves. Knowing the waves ahead (with the help of the wave lag defined in Fig.~\ref{fig:dataset}) would be very helpful for motion prediction. In practice, future waves can be obtained by a wave buoy placed in front of the platform. In this example, we fed the motion and waves together into the LSTM neural network to generate the motion prediction in heave and surge directions. 

As a starting point, we used a 60-point time window with a wave lag of 20 points to generate heave and surge motion predictions of 20 points into future. Given that the sampling rate was 1.29 Hz in full scale (10 Hz in model scale), we approximately predicted 15.5 s of motion into future. One step forward means 0.775 s into future.

\begin{figure}[ht]
	\centering
	\includegraphics[width=0.9\linewidth]{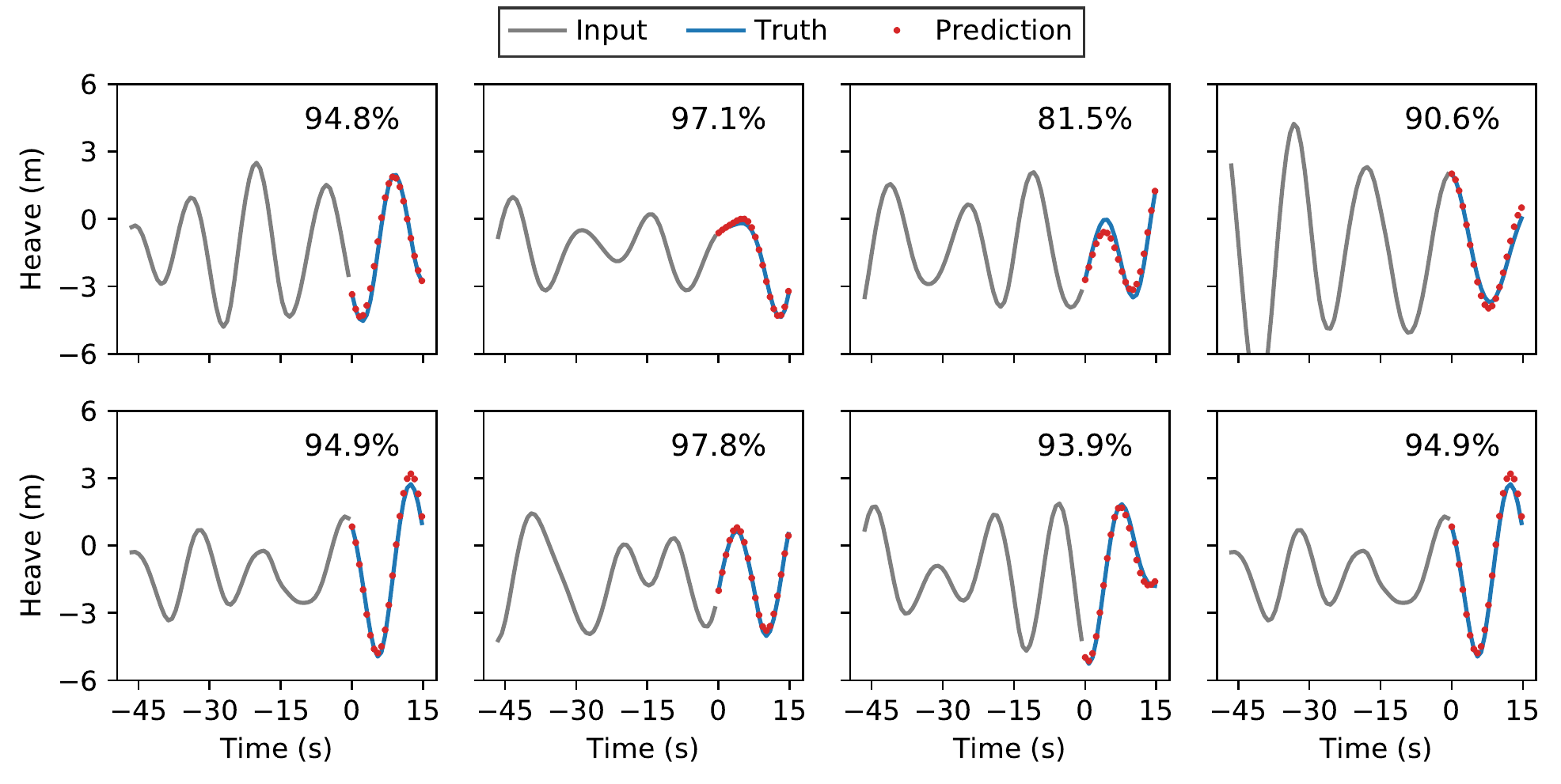}
	\caption{Examples of heave motion predictions from the proposed learning machine compared to the ground truth in the test dataset with WC2 ($H_s=13.4$ m, $T_p=14.7$ s, seed 1).}
	\label{fig:ex1-heave-base}
\end{figure}

\begin{figure}[ht]
	\centering
	\includegraphics[width=0.9\linewidth]{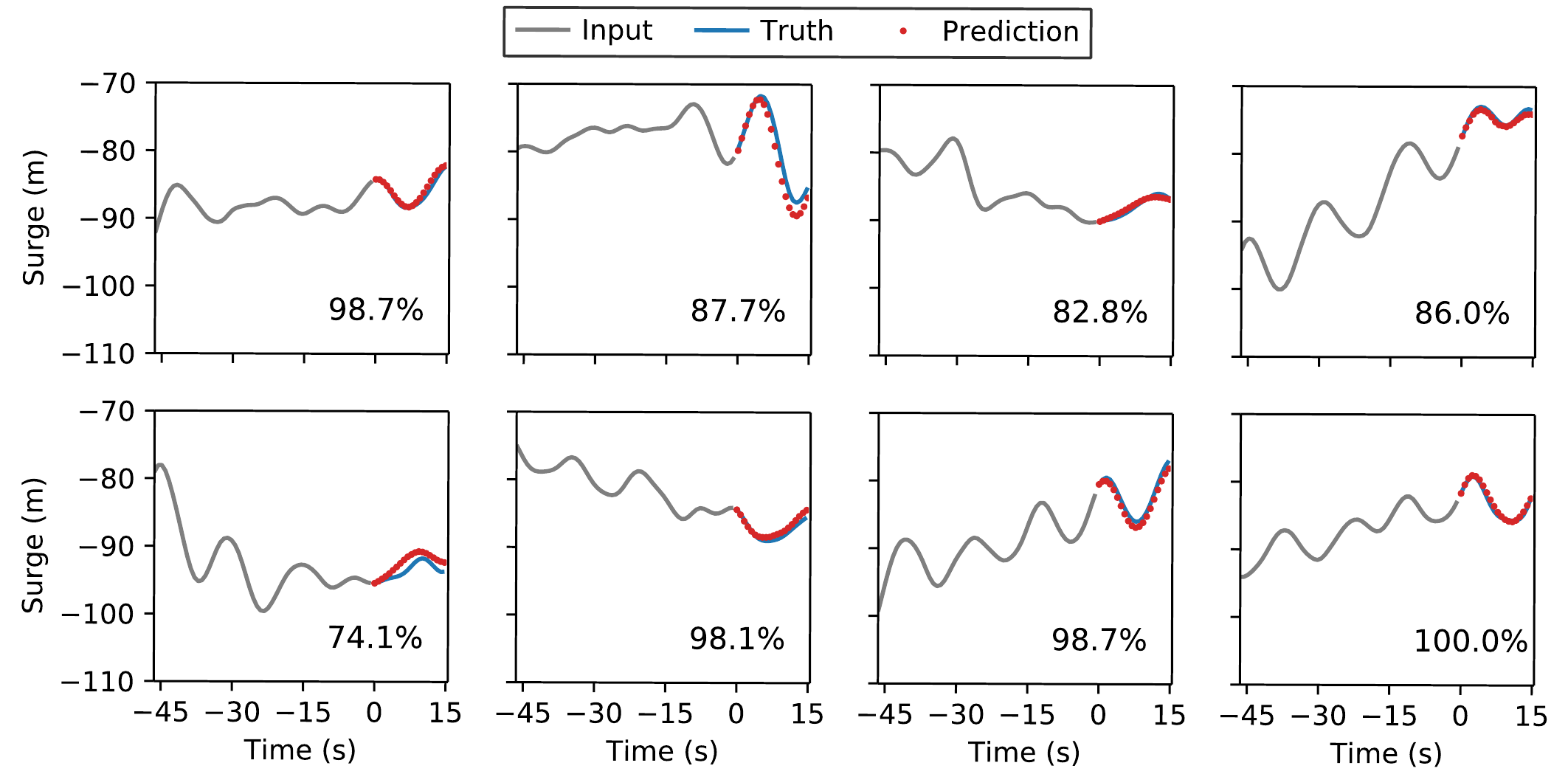}
	\caption{Examples of surge motion predictions from the proposed learning machine compared to the ground truth in the test dataset with WC2 ($H_s=13.4$ m, $T_p=14.7$ s, seed 1).}
	\label{fig:ex1-surge-base}
\end{figure}

Figures~\ref{fig:ex1-heave-base} and \ref{fig:ex1-surge-base} show some predictions from the present ML model on the test dataset. Although similar wave conditions were used for model training, the test prediction was made on unseen wave conditions. The prediction agrees very well with the ground truth. For heave motion, the prediction successfully follows the wave-frequency oscillations. For surge motion, not only the wave-frequency component but also the slow-drift motion were predicted.

We define the prediction accuracy as follows:
\begin{equation}
	\text{Acc}=1-\Big|1-\frac{\text{Area}\big(\mathscr{F}(\mathbf{X}_p)-\text{Mean}(\mathscr{F}(\mathbf{X}_p))\big)}{\text{Area}\big(\mathbf{Y}_p-\text{Mean}(\mathbf{Y}_p)\big)}\Big|,
\end{equation}
where $\text{Mean}(.)$ is the mean, $\text{Area}(.)$ is the area under the curve, which is numerically obtained by a trapezoidal rule. $\text{Acc}$ is always less than 1, and values closer to 1 mean better prediction. The average accuracy was close to 90\% in the aforementioned prediction examples (see Figs.~\ref{fig:ex1-heave-base} and \ref{fig:ex1-surge-base}).

\begin{figure}[ht]
	\centering
	\begin{subfigure}{0.4\linewidth}
         \centering
         \includegraphics[width=\linewidth]{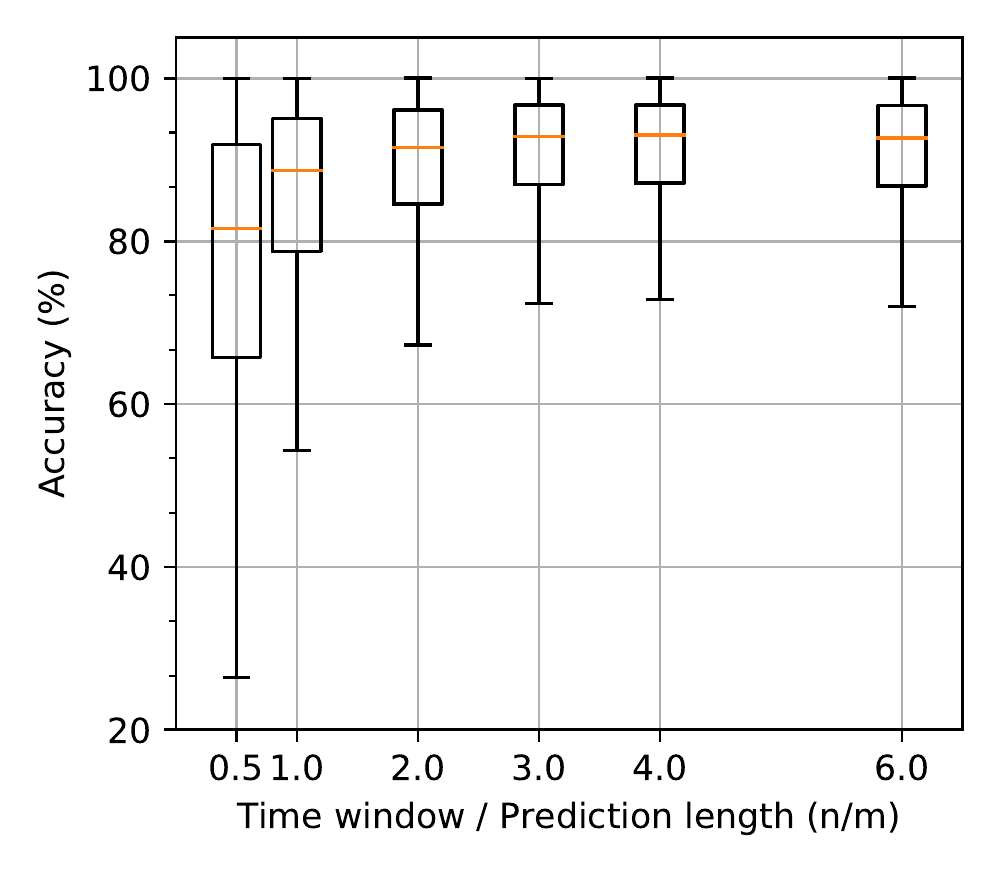}
         \caption{Heave}
	\end{subfigure}
	\begin{subfigure}{0.4\linewidth}
		\centering
		\includegraphics[width=\linewidth]{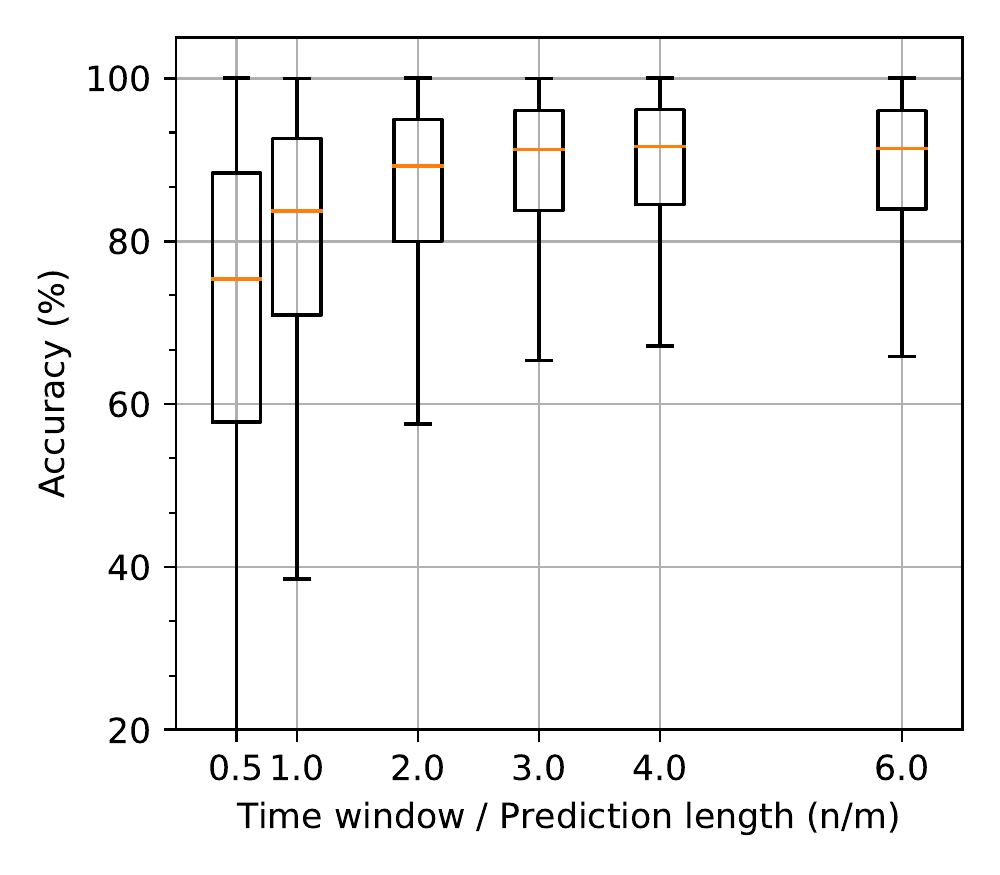}
		\caption{Surge}
    \end{subfigure}
	\caption{Statistical boxplots of prediction accuracy on the test dataset with varying time window $n$ for (a) heave motion prediction and (b) surge motion prediction.}
	\label{fig:ex1:acc:tw}
\end{figure}

For a fixed prediction length $m=20$ and wave lag $w=20$, we changed the time window $n$ from 10 to 120 in the training process. As shown in Figure~\ref{fig:ex1:acc:tw}, for both heave and surge motions, a larger time window means better prediction results. Given that the LSTM cells share the parameters in all cells (see Fig.~\ref{fig:cells:lstm}), more data used for prediction do not enlarge the parameter space. The training time barely changed with different time windows. In practice, we recommend that the time window should be at least three times the prediction length ($n \geq 3m$). Generally, in terms of accuracy, the heave motion prediction is better than that of surge.

\begin{figure}[ht]
	\centering
	\begin{subfigure}{0.4\linewidth}
         \centering
         \includegraphics[width=\linewidth]{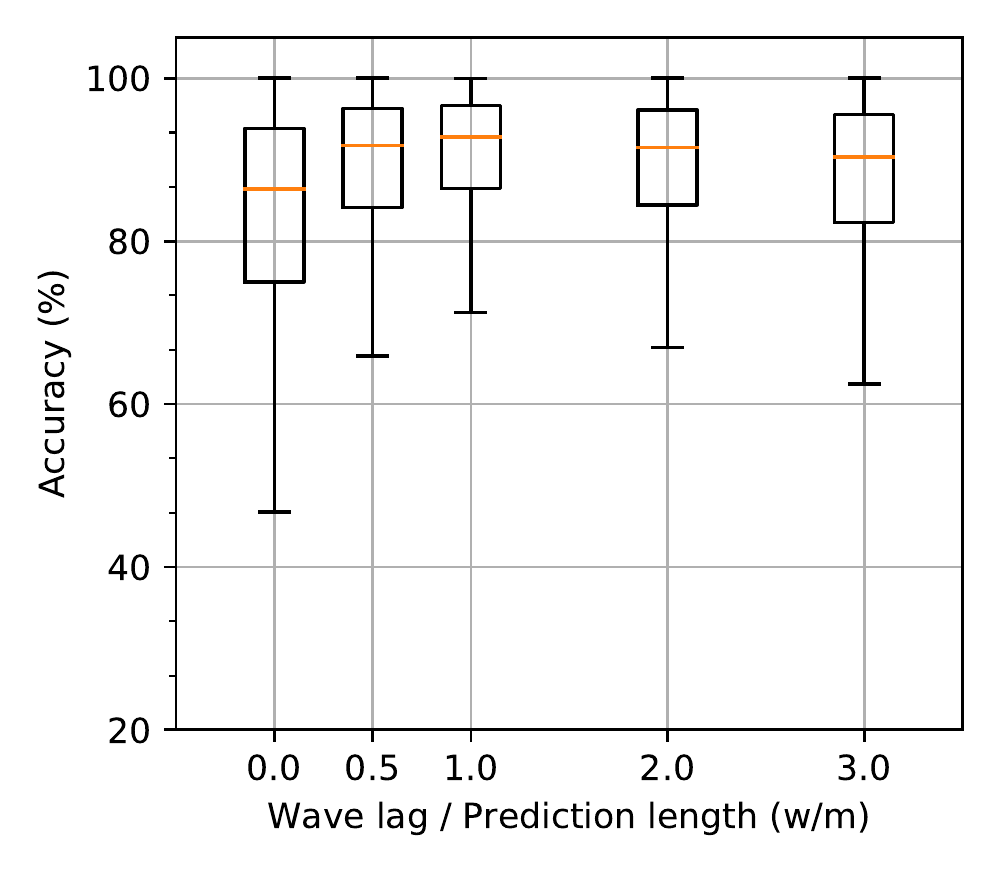}
         \caption{Heave}
	\end{subfigure}
	\begin{subfigure}{0.4\linewidth}
		\centering
		\includegraphics[width=\linewidth]{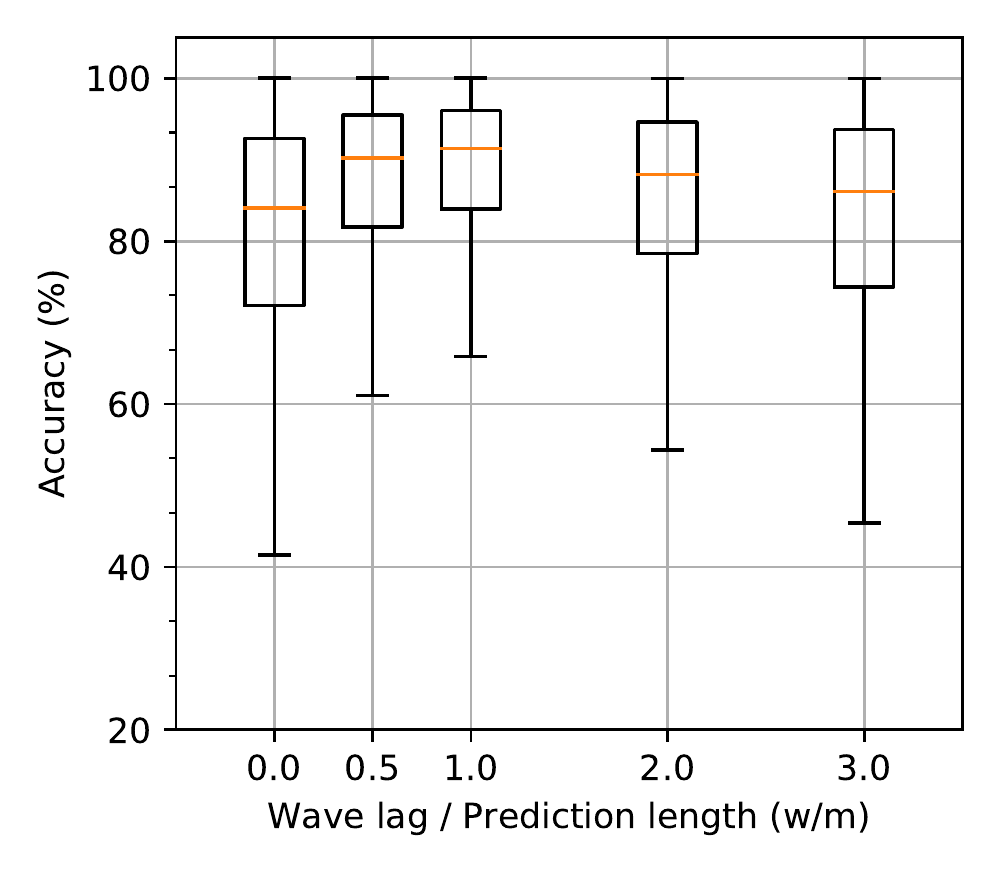}
		\caption{Surge}
    \end{subfigure}
	\caption{Statistical boxplots of prediction accuracy on the test dataset with varying wave lag $w$ for (a) heave motion prediction and (b) surge motion prediction.}
	\label{fig:ex1:acc:wl}
\end{figure}

Another hyperparameter is the wave lag, which is defined as the number of points of incident waves we could know ahead. We fixed the prediction length and time window as $m=20$ and $n=60$. Then, we changed the wave lag from 0 to 60. The accuracy boxplots are shown in Fig.~\ref{fig:ex1:acc:wl}. It is seen that when the wave lag was equal to the prediction length we obtained the best prediction. When the wave lag was larger than the prediction length, more future waves were known. However, the most related wave elevations required more steps to go through the LSTM cells. Some of the information could not be effectively used for prediction. The accuracy decreased with increasing wave lag when $w > m$.

\begin{figure}[ht]
	\centering
	\includegraphics[width=0.9\linewidth]{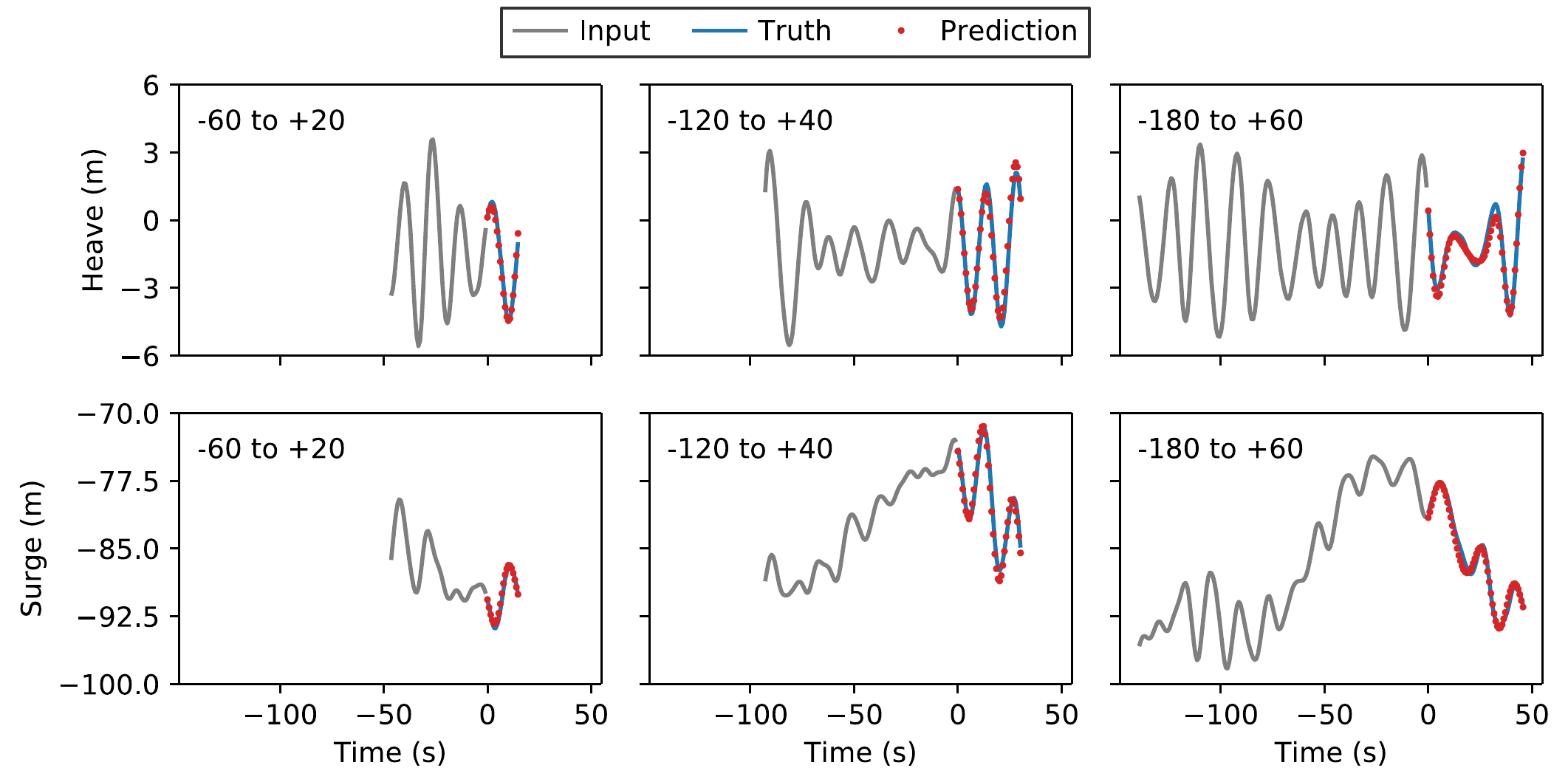}
	\caption{Examples of surge and heave motion predictions from the proposed learning machine compared to the ground truth with different prediction lengths.}
	\label{fig:ex1-predictionlength}
\end{figure}

We kept the ratio of $n/m$ and $w/m$ as 3 and 1, respectively, and then extended the prediction length ($m$) up to 60 points (approximately 46.5 s). The results are shown in Fig.~\ref{fig:ex1-predictionlength}. The trained model could predict the surge and heave motions very well. In particular, concerning surge motion, the slow-drift motion was successfully predicted. We believe that the prediction can be further extended. For the present LSTM model, extending the prediction length would not enlarge the parameter space, which means that the training time would still be under control. This is an important advantage of the proposed LSTM model for motion prediction.

In summary, we can draw the following conclusions:
\begin{itemize}
	\item The best choice of time window is three times the prediction length ($n=3m$).
	\item The best choice of wave lag is to set the same value as the prediction length ($w=m$).
	\item The prediction length can be extended from 46.5 s (three wave cycles) on if we could know the corresponding length of future waves.
\end{itemize}

\subsection{Example 2: prediction based on noisy measurements}
\begin{figure}[ht]
	\centering
	\includegraphics[width=0.9\linewidth]{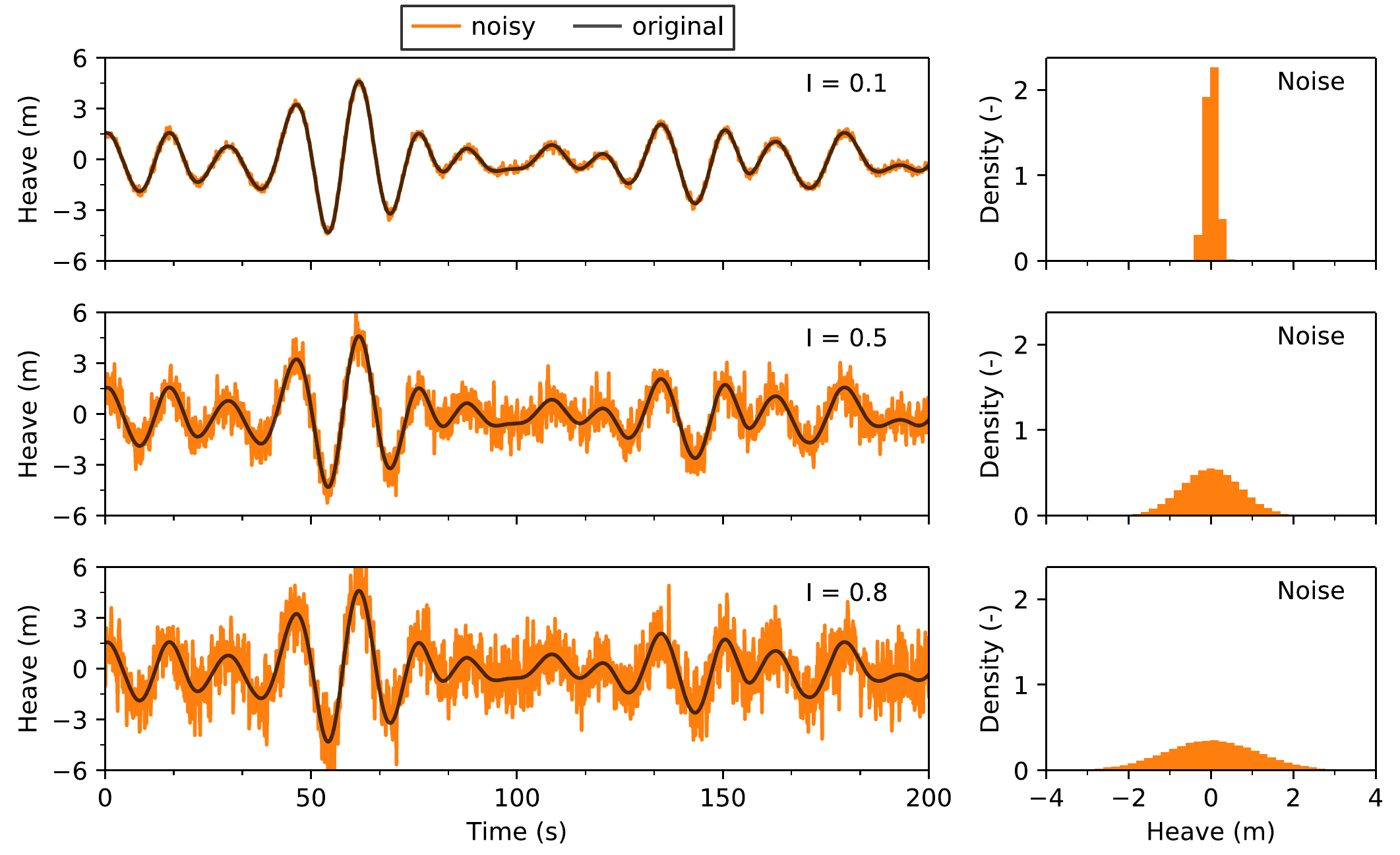}
	\caption{Time series of heave motions with different noisy level $I$ (WC5, 1000 yr cyclone, short Tp)}
	\label{fig:ex2:noisyinput}
\end{figure}

In practice, measured data always contain noise. As a further step, the effect of noise was investigated through the example described next. The noisy input is denoted as $\{x(t_i)+n(t_i), i\in N\}$, where $\{n(t_i)\}$ follows a Gaussian distribution.
\begin{equation}
	p(n(t_i))=\frac{1}{\sqrt{2\pi(I\sigma_i)^2}}\text{e}^{-\frac{n^2}{2(I\sigma_i)^2}},
\end{equation} \label{equ:ex2:i}
where $\sigma_i$ is the standard deviation of the whole 30-min time series $\{x(t_i)\}$, and $I$ is the level of noise. Figure~\ref{fig:ex2:noisyinput} shows examples of the input heave motion with different levels of noise.

\begin{figure}[ht]
	\centering
	\includegraphics[width=0.4\linewidth]{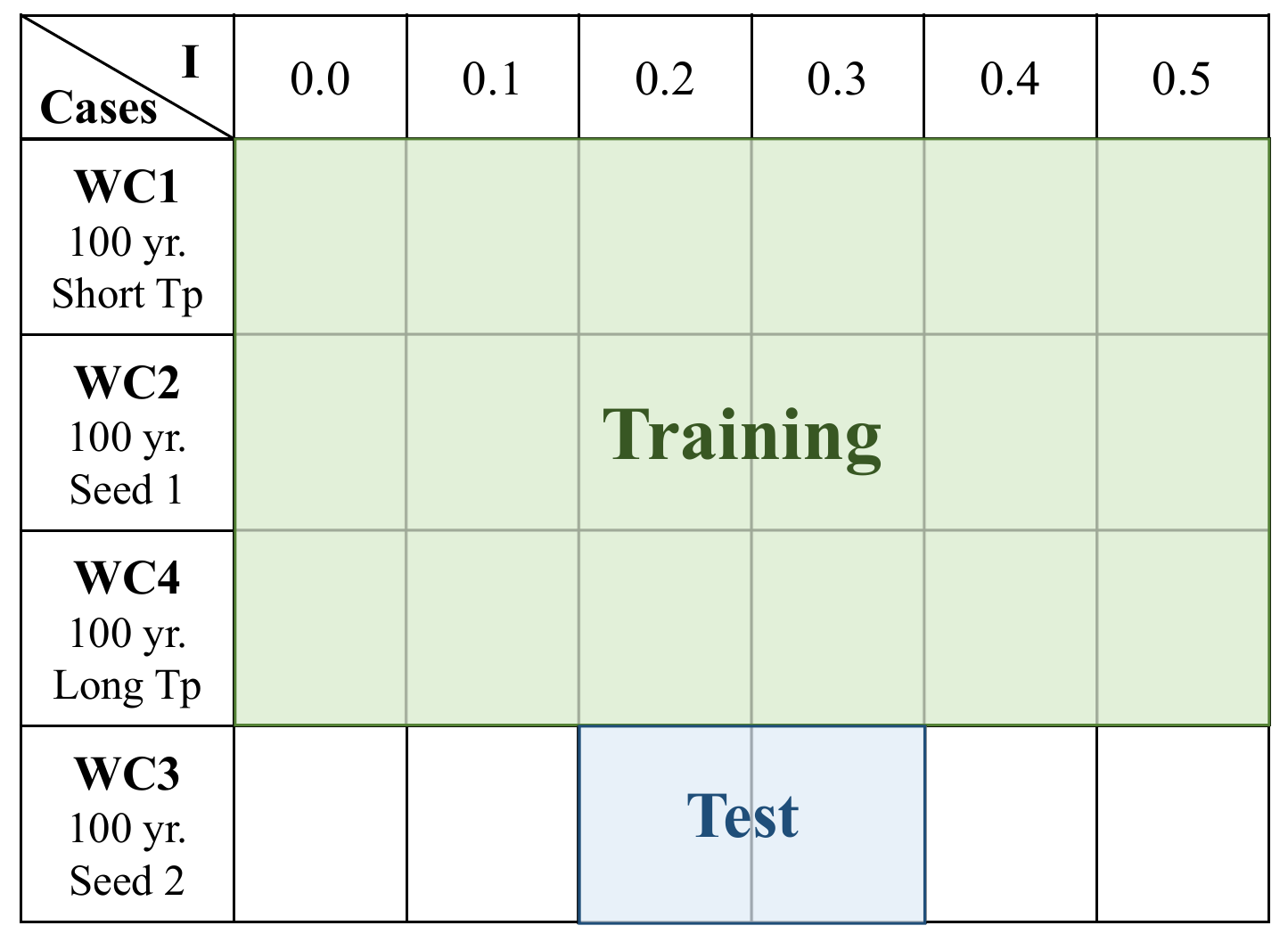}
	\caption{Illustration of the training and test datasets for Example 2.}
	\label{fig:ex2:dataset}
\end{figure}

In this case, the training dataset encompassed three cases with different wave conditions (WC1, WC3, and WC4); WC2 was still used for test. For each wave case, the original data were combined with different levels of noise to enlarge the training and test datasets. The composition of the dataset is illustrated in Fig.~\ref{fig:ex2:dataset}.

In this example, we used noisy waves and motions (with the same noise level) as input, and the original smooth data as ground truth to train the LSTM model. The prediction length was 40, and the corresponding time window and wave lag were set to 120 and 40, respectively. The same level of noise was added to both the motion and wave inputs.

\begin{figure}[ht]
	\centering
	\includegraphics[width=0.9\linewidth]{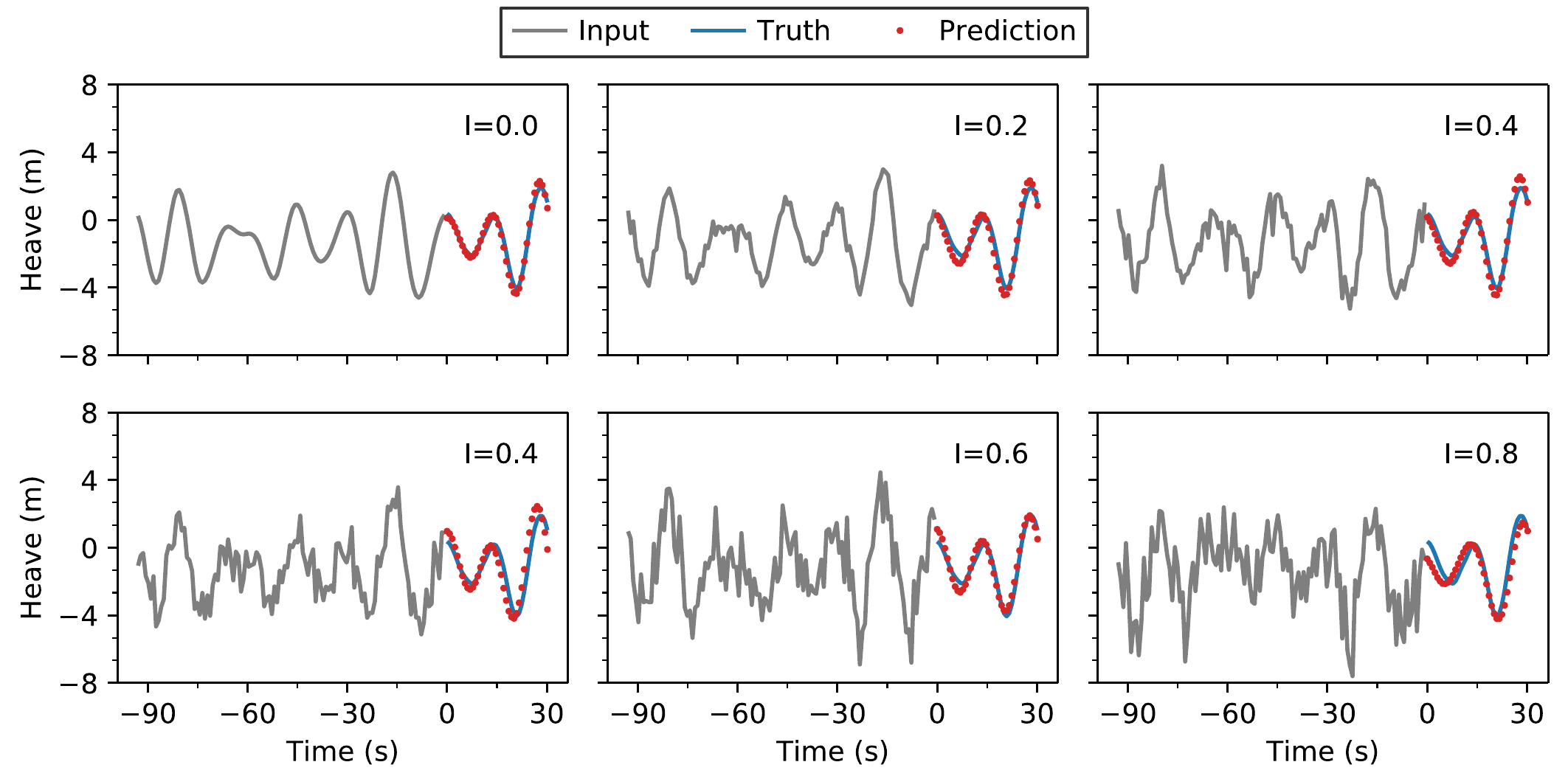}
	\caption{Heave motion predictions with different levels of noise on WC3 (100 yr Seed2).}
	\label{fig:ex2:nl:heave}
\end{figure}

\begin{figure}[ht]
	\centering
	\includegraphics[width=0.9\linewidth]{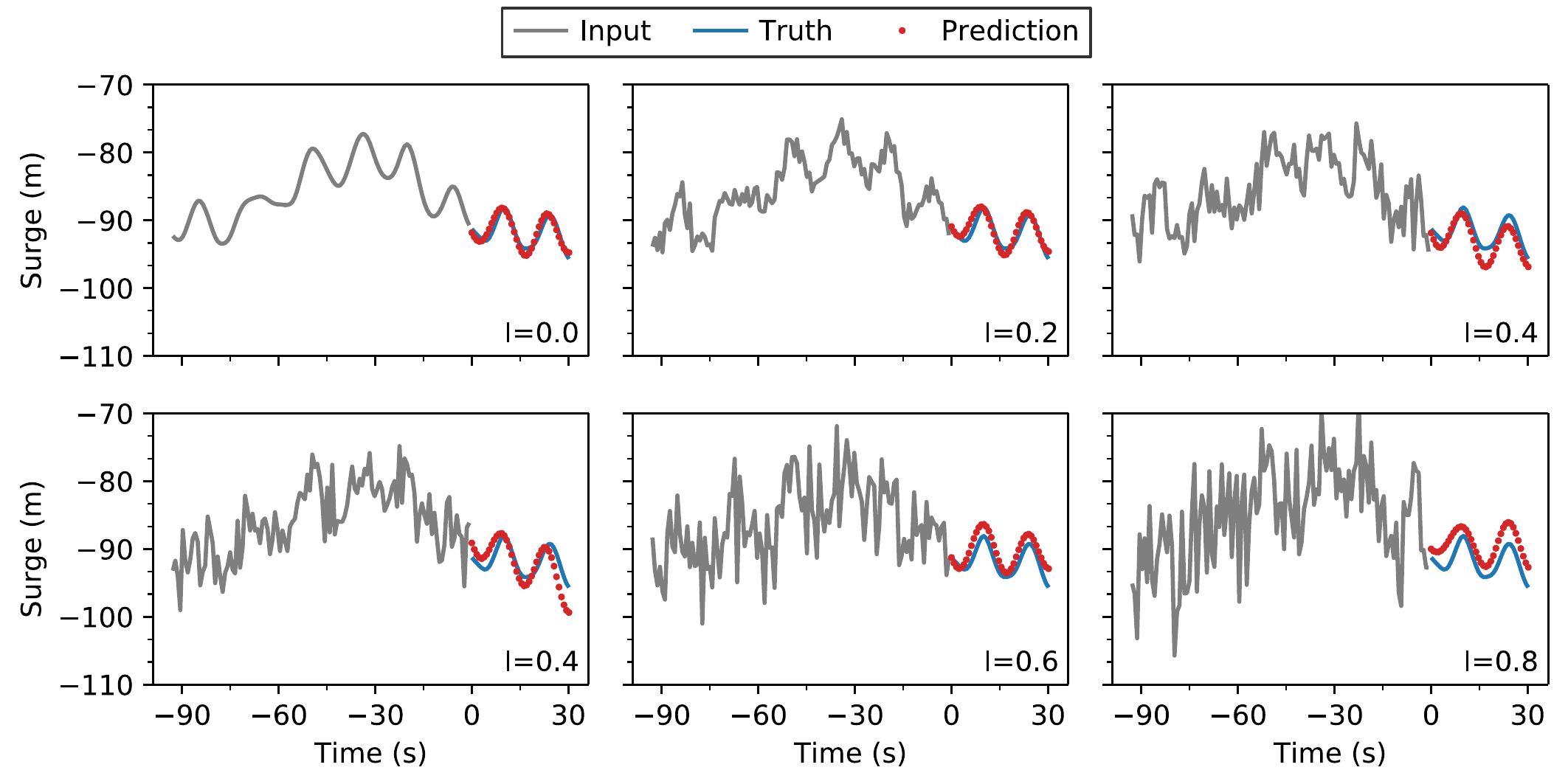}
	\caption{Surge motion predictions with different levels of noise on WC3 (100 yr Seed2).}
	\label{fig:ex2:nl:surge}
\end{figure}

Figures~\ref{fig:ex2:nl:heave} and \ref{fig:ex2:nl:surge} show the heave and surge motion predictions on the test dataset with different levels of noise. Note that the learning machine effectively removed noise and made proper motion prediction. In general, the accuracy of heave prediction was better than that of surge prediction. The prediction accuracy decreased with a higher level of noise. Note also that the surge motion contained wave-frequency and slow-drift components. With a high level of noise ($I>0.6$), it is notably difficult to distinguish the original motion from the noise. Therefore, in this situation the surge motion prediction was worse. 

During the training process, we only used noise levels ranging from 0 to 0.5. However, the model effectively worked on higher noise levels up to 0.8. In the present model, three FC layers with 50 neurons were used after the LSTM layers. These layers act as an encoder-decoder process. The key features of the motions are extracted and stored in the first FC layer (FC 1). Then, the predicted time series of the motion are obtained by recombination of these key features by serval FC layers. From this point of view, artificial noise can help the model to understand which information is important and must be stored as key features to represent the time series of motion. Once the training process is finished, the model can effectively remove noise from the measured data and then generate the proper motion prediction. In other words, the prediction model can address uncertainties of the wave and motion measurements.

\subsection{Example 3: prediction only based on motion }
In the above two examples, we predicted the motions with the help of waves. In practice, wave measurement is not an easy task. Therefore, we developed a model to make prediction only based on the motion itself. As shown in Fig.~\ref{fig:ex1:acc:wl}, when the wave lag was zero, the prediction accuracy dropped significantly. It means that the architecture of the model must be changed to improve the performance. 
In this case, the problem can be formulated as follows: given
\begin{align}
	&\mathbf{X}^{n \times 1}_p = [x(t_{p-n}), x(t_{p-n}), \cdots, x(t_{p-1})]^T, \\
	&\mathbf{Y}^{m \times 1}_p = [x(t_{p}), x(t_{p+1}), \cdots, x(t_{p+m-1})]^T,
\end{align}
find the learning machine,
\begin{equation}
	\mathscr{F}(\mathbf{x}; \mathbf{p})\approx \mathbf{y}.
\end{equation}

\begin{figure}[ht]
	\centering
	\includegraphics[width=0.7\linewidth]{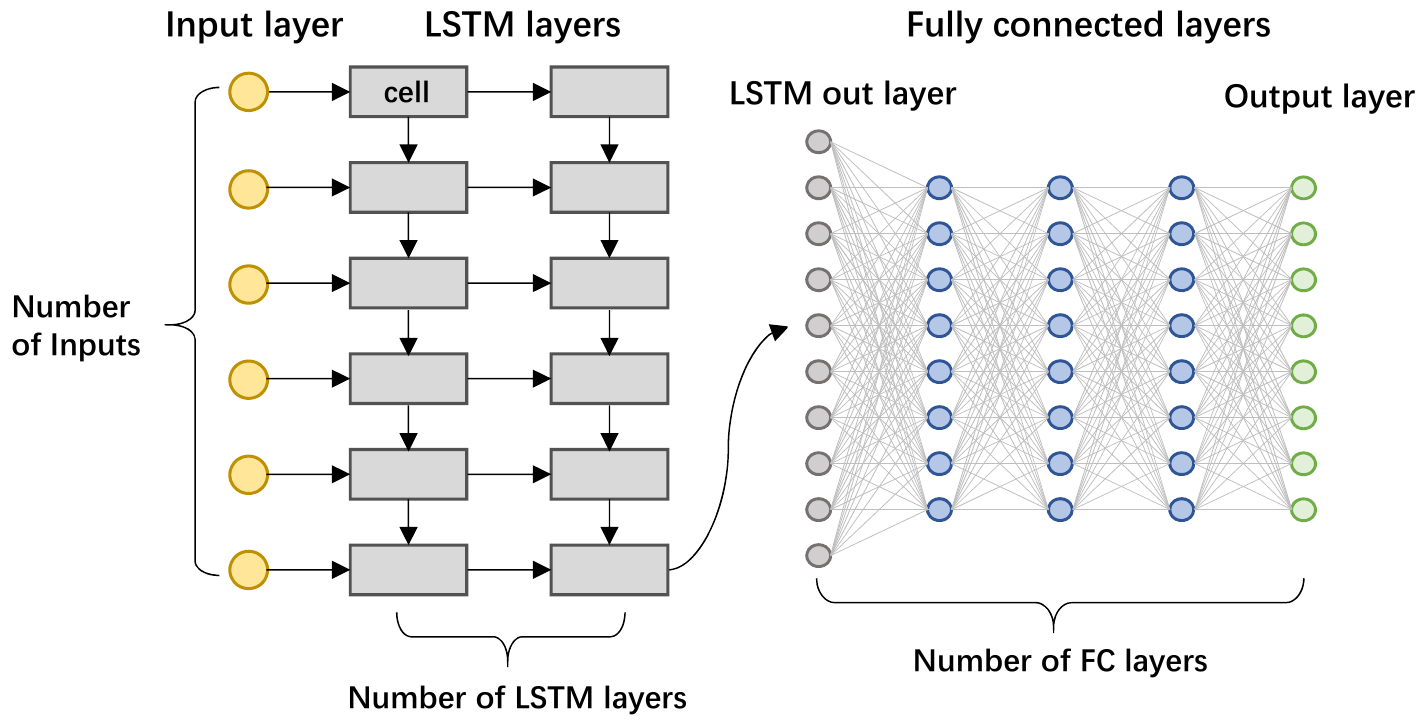}
	\caption{Illustration of the ML model structure.}
	\label{fig:ex3-mdl}
\end{figure}

The basic model structure consists of layers and neurons in LSTM and FC layers, as shown in Fig.~\ref{fig:ex3-mdl}. The previous setting of the model structure can be found in Tab.~\ref{tab:basicmodel}.

\begin{figure}[ht]
	\centering
	\includegraphics[width=0.75\linewidth]{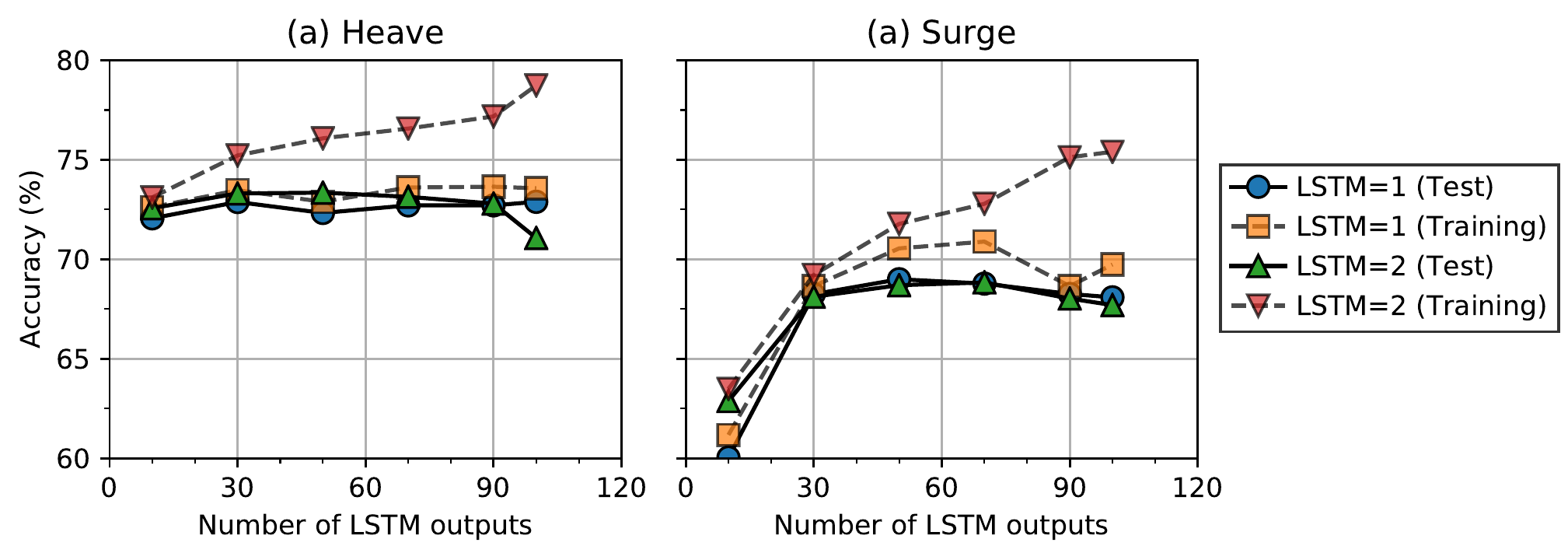}
	\caption{Average accuracy of (a) heave and (b) surge motion predictions on both the training and test datasets with different LSTM layer structures}
	\label{fig:ex3-hc}
\end{figure}

\begin{figure}[ht]
	\centering
	\includegraphics[width=0.75\linewidth]{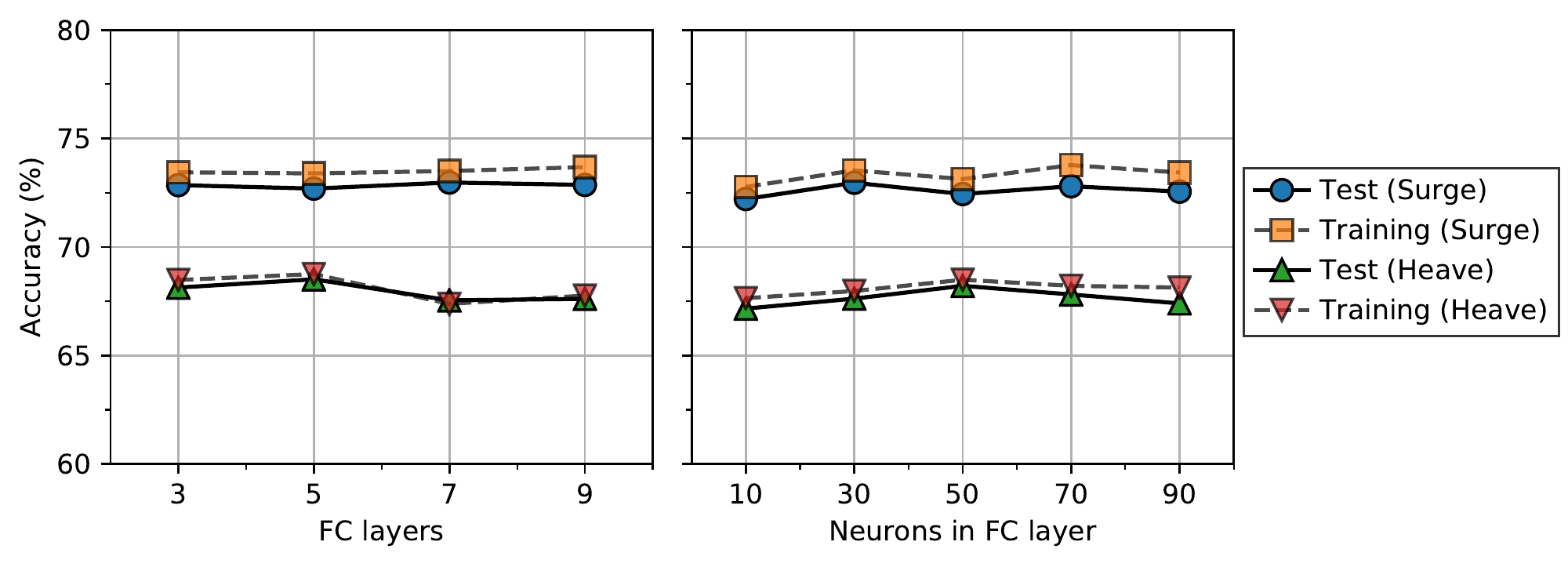}
	\caption{Average accuracy of heave and surge motion predictions on both the training and test datasets with different FC layers.}
	\label{fig:ex3-fc}
\end{figure}

We set the prediction length to 20 and the corresponding time window to 60. Figs.~\ref{fig:ex3-hc} and \ref{fig:ex3-fc} show the average accuracy of the prediction on both the training and test datasets when the structure of the model was changed. Note that the training and test prediction accuracies constitute a critical criterion to determine whether under- or over-fitting is occurring. If the model performs much better on the training set than on the test set, overfitting is likely taking place. In Fig.~\ref{fig:ex3-hc}, we increased the number of LSTM outputs and tried to add an extra LSTM layer. This extra layer did not improve the prediction on the test set but introduced significant overfitting when the LSTM output was larger than 60. For surge motion prediction, the LSTM outputs should be larger than 30 to obtain reasonable results. Further increase of the LSTM outputs did not improve the prediction accuracy. Therefore, we conclude that the best choice is one LSTM layer with 30 outputs.

The next question was how FC layers affect the results. As shown in Fig.~\ref{fig:ex3-fc}, we found that both the number of FC layers and the number of neurons in each FC layer did not change the accuracy. This is because the FC layers cannot provide any new information. The role of the FC layers is to recombine the output of the LSTM cells to obtain the final prediction. Therefore, we believe that 3 FC layers with 30 neurons in each layer are enough for the present model. 

\begin{figure}[ht]
	\centering
	\includegraphics[width=0.8\linewidth]{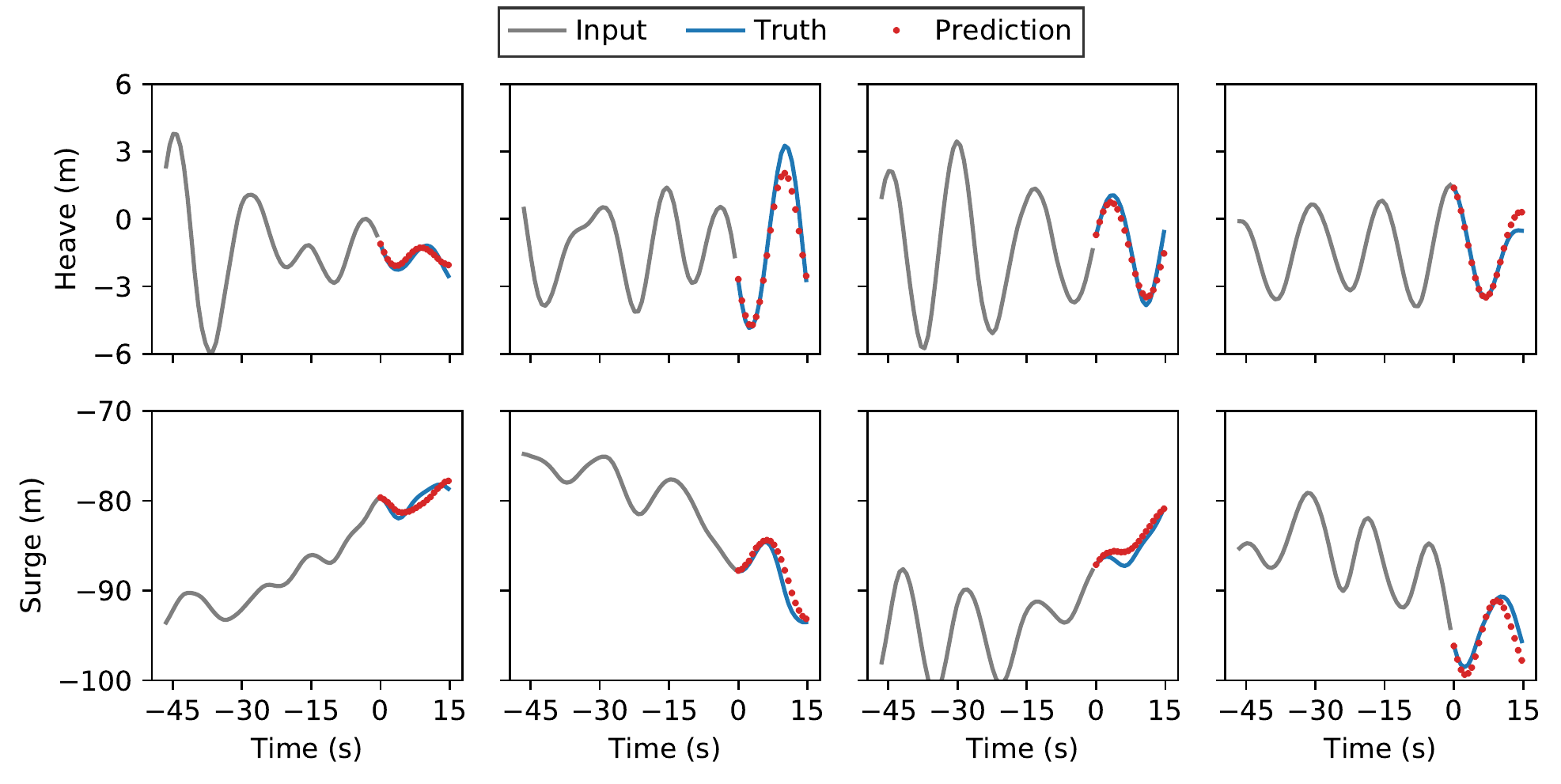}
	\caption{Examples of heave and surge motion predictions from the learning machine only based on motion.}
	\label{fig:ex3-heave}
\end{figure}

Figure~\ref{fig:ex3-heave} shows predictions from the proposed model. Note that the learning machine effectively predicted the motion into future with acceptable accuracy. Without measurements of incident waves, the average accuracy of the prediction fell approximately 10\%. 

\section{Concluding remarks}
In this study, we developed a machine learning model to predict heave and surge motions of a semi-submersible platform. The training and test datasets came from a model test carried out at SJTU deep-water ocean basin, China. The motion and measured waves were fed into LSTM cells and then went through serval FC layers to obtain the prediction results. In the first example, with the help of measured waves, we could predict the motions up to 46.5 s into future, and the average accuracy of the prediction was approximately 90\%. We propose that, to obtain the best prediction, the time window should be three times the prediction length and the wave lag should be equal to the prediction length. Then, the model was used for noisy inputs. With the noise-extended dataset, the trained model could effectively make predictions with noise levels up to 0.8 of the input data. In the last example, a more complex model was developed for prediction only based on the motion itself. We found that too many FC layers are not necessary, and one more LSTM layer gives rise to overfitting. The average prediction accuracy fell approximately 10\% to 15\% compared with the accuracy obtained with the help of wave measurements.

The present LSTM model demonstrates the strong ability of ML models to predict vessel wave-excited motions. To obtain better predictions, a more complex model and an enlarged dataset should be used. The architecture of the model is crucial and for many applications, determining such architecture is an art that relies on personal ability to balance the trade-off between prediction accuracy and computation cost. In this study, we provide some guidelines for model construction and try to reveal the physical meaning of each part of the model. Whether the proposed model architecture is valid for other type of offshore structures, e.g., spar, TLP, or FPSO , is the next question to be addressed.

\section{Acknowledgements}
This study is supported by the National Natural Science Foundation of China (Grant No. 51679137), Shanghai Sailing Program (Grant No. 20YF1419700) and State Key Laboratory of Ocean Engineering (Shanghai Jiao Tong University) (Grant No. 1915).

\bibliographystyle{unsrt}


\end{document}